%% file: acl_latex.tex
\pdfoutput=1

\documentclass[11pt]{article}

\usepackage[]{acl}

\usepackage{times}
\usepackage{latexsym}

\usepackage[T1]{fontenc}

\usepackage[utf8]{inputenc}

\usepackage{microtype}

\usepackage{multirow}

\usepackage{amsmath,amsthm,mathtools}
\usepackage{subcaption}
\usepackage{graphicx}
\usepackage{float}
\usepackage{colortbl}
\usepackage[most]{tcolorbox}
\usepackage{tabularx}
\newcolumntype{Y}{>{\centering\arraybackslash}X}
\newcolumntype{M}[1]{>{\centering\arraybackslash}m{#1}} 

\usepackage{CJKutf8} 
\usepackage{pgfplots}
\pgfplotsset{compat=1.18}
\definecolor{color1}{RGB}{9,147,150}
\definecolor{color4}{RGB}{0,18,25}
\definecolor{color3}{RGB}{238,155,0}
\definecolor{color2}{RGB}{174,32,18}

\title{Word Alignment as Preference for Machine Translation}

\author{Qiyu Wu$^{1}$,
        Masaaki Nagata$^{2}$,
        Zhongtao Miao$^{1}$,
        Yoshimasa Tsuruoka$^{1}$\\
$^{1}$\normalfont{The University of Tokyo}, Tokyo, Japan \\
$^{2}$\normalfont{NTT Communication Science Laboratories, NTT Corporation, Kyoto, Japan} \\
$^{1}$\normalfont{\texttt{\{qiyuw, mzt, yoshimasa-tsuruoka\}@g.ecc.u-tokyo.ac.jp}} \\
$^{2}$\normalfont{\texttt{masaaki.nagata@ntt.com}}
}

\newcommand{\methodname}{WAP}
\newcommand{\hao}{hallucination and omission}
\newcommand{\hoo}{hallucination or omission}

\newcommand{\redmark}[1]{\textcolor{red}{\textless{}\textless{}\textless{}#1\textgreater{}\textgreater{}\textgreater{}}}
\newcommand{\bluemark}[1]{\textcolor{blue}{\textless{}\textless{}\textless{}#1\textgreater{}\textgreater{}\textgreater{}}}
\newcommand{\chinese}[1]{\begin{CJK*}{UTF8}{gbsn}#1\end{CJK*}}

\begin{document}
\maketitle

\begin{abstract}
The problem of \hao{}, a long-standing problem in machine translation (MT), is more pronounced when a large language model (LLM) is used in MT because an LLM itself is susceptible to these phenomena.
In this work, we mitigate the problem in an LLM-based MT model by guiding it to better word alignment.
We first study the correlation between word alignment and the phenomena of \hao{} in MT. Then we propose to utilize word alignment as preference to optimize the LLM-based MT model.
The preference data are constructed by selecting chosen and rejected translations from multiple MT tools. Subsequently, direct preference optimization is used to optimize the LLM-based model towards the preference signal. 
Given the absence of evaluators specifically designed for \hao{} in MT, we further propose selecting hard instances and utilizing GPT-4 to directly evaluate the performance of the models in mitigating these issues. We verify the rationality of these designed evaluation methods by experiments, followed by extensive results demonstrating the effectiveness of word alignment-based preference optimization to mitigate \hao.
On the other hand, although it shows promise in mitigating \hao{}, the overall performance of MT in different language directions remains mixed, with slight increases in BLEU and decreases in COMET.

\end{abstract}

\input{sections/intro}
\input{sections/related}

\input{sections/method}
\input{sections/exp}

\section{Conclusion}
The problem of \hao{}, a long-standing problem in MT, could become more severe when an LLM is used because an LLM itself could hallucinate or omit in nature.
In this paper, our aim is to mitigate this problem in LLM-based MT by optimizing the model toward a preference for better word alignment. We construct preference datasets by collecting translations using multiple MT tools and selecting the preference pair with a higher coverage score output by a word aligner. DPO is then utilized to optimize the model towards the word-aligned preference. As evaluation of \hao{} is challenging, we design experiments that include selecting hard instances and using GPT-4 to directly predict coverage score, ensuring an effective evaluation, which indicates that the proposed \methodname{} mitigates \hao{}, especially in hard instances. On the other hand, although \methodname{} shows promise in mitigating \hao{}, the overall performance of MT in different language directions remains mixed, with slight increases in BLEU and decreases in COMET.

\section*{Limitation}
The first limitation of our method stems from the imperfections of the word alignment model. Within our approach, it is inevitable to encounter some alignment errors, which we address through a filtering method. However, this solution adds complexity and clutter to the method.
Additionally, the effectiveness of our method is diminished for low-resource language translations due to the limited number of parallel sentences available. From the perspective of experiments, we only evaluate the methods in English-centric translation pairs due to the lack of Non-English data, in which \hao{} could happen more frequently. In particular, the WMT-2024 General Machine Translation Task~\cite{kocmi2024preliminary} has adopted non-English language pairs, such as Czech-to-Ukrainian and Japanese-to-Chinese, which could expand our work in the future.
Moreover, our reliance on the GPT-4 API to evaluate the results introduces a significant cost factor. In future work, our objective is to find a cost-free alternative to this evaluation process.
Lastly, although \methodname{} shows promise in mitigating \hao{}, the overall performance of MT in different language directions remains mixed, with slight increases in BLEU and decreases in COMET.

\section*{Ethical Statement}
All datasets and checkpoints used in this paper are copyright-free for research purposes. Previous studies are properly cited and discussed. This research aims to improve LLM-based machine translation models with word alignment preference data, and the preference is made by an automatic word aligner. We do not introduce additional bias to particular communities. We have obtained the consent of the annotation volunteers for this study.


\bibliography{anthology,custom}
\bibliographystyle{acl_natbib}

\clearpage
\appendix 
\input{sections/appendix}



\end{document}

%% file: sections/intro.tex
\section{Introduction}
Large language models (LLMs) have been evolving rapidly and showing predominant performance in many natural language processing (NLP) tasks~\citep{gpt3, achiam2023gpt, touvron2023llama}.
However, in machine translation (MT), the use of decoder-only LLMs is still limited due to issues such as model size~\citep{alma} and low-resource languages~\citep{gpt-mt-2023}. Conventional encoder-decoder MT models trained on parallel corpora still dominate in practice~\citep{nllb2022}.
One of the primary concerns of applying an LLM to MT is reliability. Although it does not happen frequently, an LLM is known to hallucinate \citep{Dhuliawala2023ChainofVerificationRH, Zhang2023SirensSI, Bang2023AMM} as it is pre-trained to predict the next token in very large-scale raw texts.
Specifically in MT, LLM-based translation systems therefore could have the phenomena of \hao{}, which is also a long-term challenge in the field of MT \citep{yang-etal-2019-reducing, vamvas-sennrich-2022-little}, known as over- and under-translation.
In particular, in the very recent WMT-2024 General Machine Translation Task~\cite{kocmi2024preliminary}, a newly released LLM-based MT model Unbabel Tower~\cite{tower_llm_2024} has achieved the highest accuracy in most language pairs, demonstrating the promise of LLM in MT, but also showing the significance of the problem of \hao{}.
As a result, we attempt to mitigate the \hao{} in LLM-based MT to improve its practicality in this work. 

\begin{figure}[!t]
    \centering
    \begin{subfigure}{\linewidth}
         \centering
         \includegraphics[width=\linewidth]{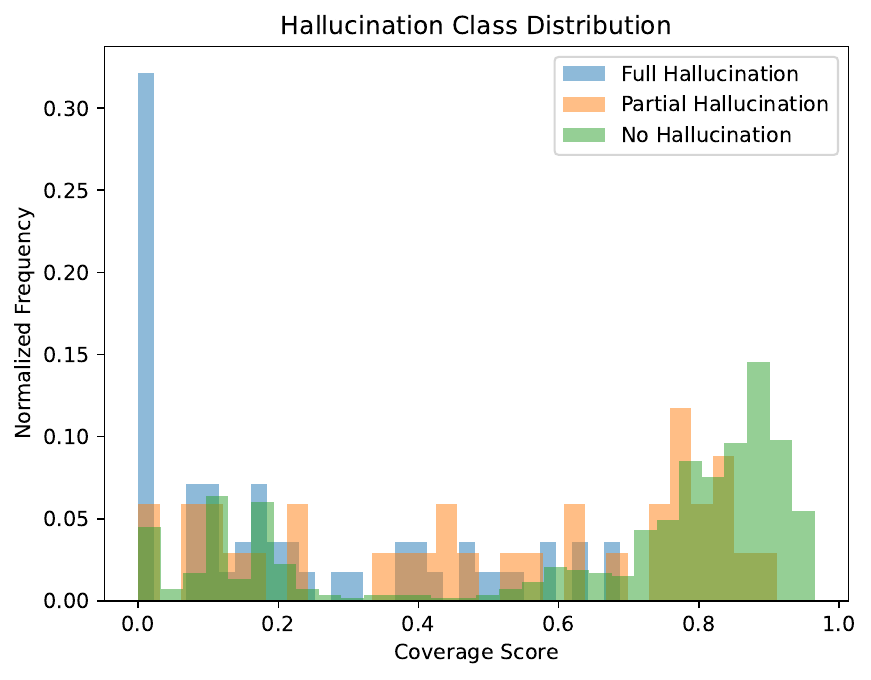}
         \caption{Coverage distribution of different hallucination degree.}
         \label{fig:hal}
     \end{subfigure}
     
     \begin{subfigure}{\linewidth}
         \centering
          \includegraphics[width=\linewidth]{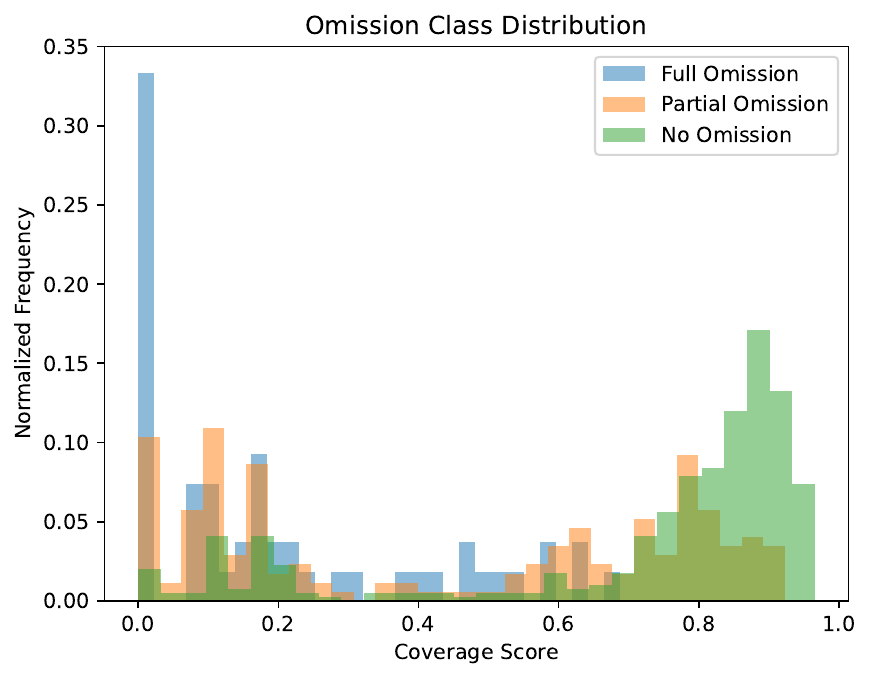}
         \caption{Coverage distribution of different omission degree.}
         \label{fig:omi}
     \end{subfigure}
     
    \caption{A preliminary experiment shows that higher coverage scores correlates to less \hao{}. The coverage scores are predicted by a word aligner~\cite{wu-etal-2023-wspalign}. The human annotation of \hao{} is from HalOmi benchmark~\cite{dale-etal-2023-halomi}. Details about the dataset and word alignment model can be found in \S\ref{sec:datasets}.}
    \label{fig:word_and_hal}
\normalsize
\end{figure}

Hallucination in MT occurs when information not present in the source text is generated in the translation, and omission occurs when some of the information in the source text is missed in the translation. As a related tool that explicitly aligns the source text and translation at the word level, word alignment is potentially positive for MT due to the nature of align and translate~\citep{BahdanauCB14}. The degree of coverage of the source text in translation could be a direct signal to identify the \hao{} in MT~\cite{Tu2016ModelingCF}.
Figure~\ref{fig:word_and_hal} shows the normalized frequency of the coverage scores predicted by a word aligner. The examples that are annotated as ``no \hoo'' tend to have a higher coverage score, while those in ``full \hoo'' are more likely to have an extremely low coverage score. ``small \hoo'' and ``partial \hoo'' distribute in the middle. As the annotations are carefully made by humans and highly correlates to the coverage scores from the word aligner, this indicates that word alignment is a simple but promising direction to mitigate these phenomena.

Consequently, we propose Word Alignment Preference (\textbf{\methodname{}}) that utilizes word alignment as a signal to optimize LLM-based MT models. \methodname{} consists of three steps: diverse translation collection, preference data construction, and preference optimization. Specifically, we collect diverse translations with multiple existing translation tools, select chosen and rejected examples with the word aligner~\citep{wu-etal-2023-wspalign}, and optimize the model on preference data using direct preference optimization (DPO)~\citep{dpo}.

Furthermore, the evaluation of \hao{} is challenging, and there is no existing evaluator specifically designed for this. Improving the BLEU and COMET score does not necessarily mean reducing \hao{} because there are other factors such as mistranslation and fluency. In addition, hallucination is relatively infrequent, although very severe once it does occur. Hence, to effectively evaluate it, we design extensive experiments that include testing on instances that potentially have the problem of \hao{}, and using GPT-4 as the evaluator with comprehensive analysis. Experimental analysis demonstrates the effectiveness of \methodname{} in mitigating \hao{} in MT.

In summary, the contributions of this work include the following:
\begin{itemize}
    \item We studied the correlation between the coverage score by word alignment and the phenomena of \hao{} in MT. From the preliminary experiments in Figure~\ref{fig:word_and_hal} we found that word alignment is a promising signal to mitigate it.
    \item In \S\ref{sec:approach} we propose a novel approach, namely \methodname{}, to construct a word alignment-based preference dataset, and use DPO to optimize the LLM-based MT model. The validity of the preference dataset is also demonstrated by direct fine-tuning on preferred and rejected translations in \S\ref{sec:ablation}.
    \item As there is no particular benchmark for evaluating the performance of MT models on \hao{}. We design various experiments, including selecting hard instances and using LLM as an evaluator in \S\ref{sec:eval_design}. The effectiveness of the evaluation, as well as the proposed \methodname{} has been validated through experiments and analysis in \S\ref{sec:exp}
\end{itemize}

%% file: sections/related.tex
\section{Related work}
\paragraph{Hallucination and omission in MT.}
Hallucinations are cases in which the model generates output that is partially or completely unrelated to the source sentence, while omissions are translations that do not include some of the input information~\citep{dale-etal-2023-halomi}. 
\citet{dale-etal-2023-detecting} explore methods that leverage the internal workings of models and external tools, such as cross-lingual sentence similarity and natural language inference models, to detect and mitigate hallucinations in MT.
HalOmi~\citep{dale-etal-2023-halomi} introduces an annotated dataset specifically designed to detect hallucinations and omissions. In Figure~\ref{fig:word_and_hal} and \S\ref{sec:eval_design} we use HalOmi as a reference to assess how these two phenomena correlate to the coverage output of the GPT-4 evaluator and the word aligner, respectively. In particular, \citet{yang-etal-2019-reducing} introduce the use of word alignment to reduce omission in MT, which partially inspires our idea.

\paragraph{Preference tuning for LLMs.}
LLMs are capable of completing tasks in the zero-shot or few-shot manner~\citep{gpt2, gpt3}. In addition, performance in downstream tasks can also be enhanced by fine-tuning them with instruction datasets~\citep{flan, flan-t5, instruct_gpt}. However, acquiring instruction datasets is costly, while obtaining preferences for LLM responses is relatively easier~\citep{dpo}.
DPO~\citep{dpo} directly optimize LLM with preference data by removing an extra reward model. We utilize DPO in this work due to the ease of use and effectiveness.
A contemporaneous preference-based method ALMA-R~\citep{cpo}, introduces contrastive preference optimization to fine-tune LLMs specifically using reference-free MT metrics and human annotation as preference. ALMA-R focuses on improving general LLM-based MT but we attempt to mitigate the \hao{} in MT. In addition, our preference data are made entirely automatically, which also draws the difference between ALMA-R and our work.
The recently released LLM-based Unbabel Tower~\cite{tower_llm_2024} has achieved the best performance in most language pairs in WMT-2024~\cite{kocmi2024preliminary}, which may complement our findings in future work.

\paragraph{Word alignment.}
Word-level information has been useful in many NLP tasks such as language pre-training~\cite{chi-etal-2021-improving, Wu2021TakingNO}, cross-lingual sentence embedding~\cite{zhang2023veco,li-etal-2023-dual,miao2024enhancing}, fine-grained visual language grounding~\cite{peng2023kosmos, wu2023towards, wu2024sga}, and particularly in word alignment for MT~\cite{BahdanauCB14,Tu2016ModelingCF}, which aligns the corresponding words in translations. 
Word aligners based on pre-trained language models \cite{JaliliSabet2020SimAlignHQ, dou2021word, nagata2020supervised,chousa-etal-2020-spanalign} have outperformed previous ones based on statistical MT~\cite{och2003systematic, dyer2013simple}. WSPAlign~\citep{wu-etal-2023-wspalign} is a pre-trained word aligner that outperforms most of the previous ones; hence we use it in the experiments.

%% file: sections/method.tex
\begin{figure*}[ht]
    \centering
    \includegraphics[width=\textwidth]{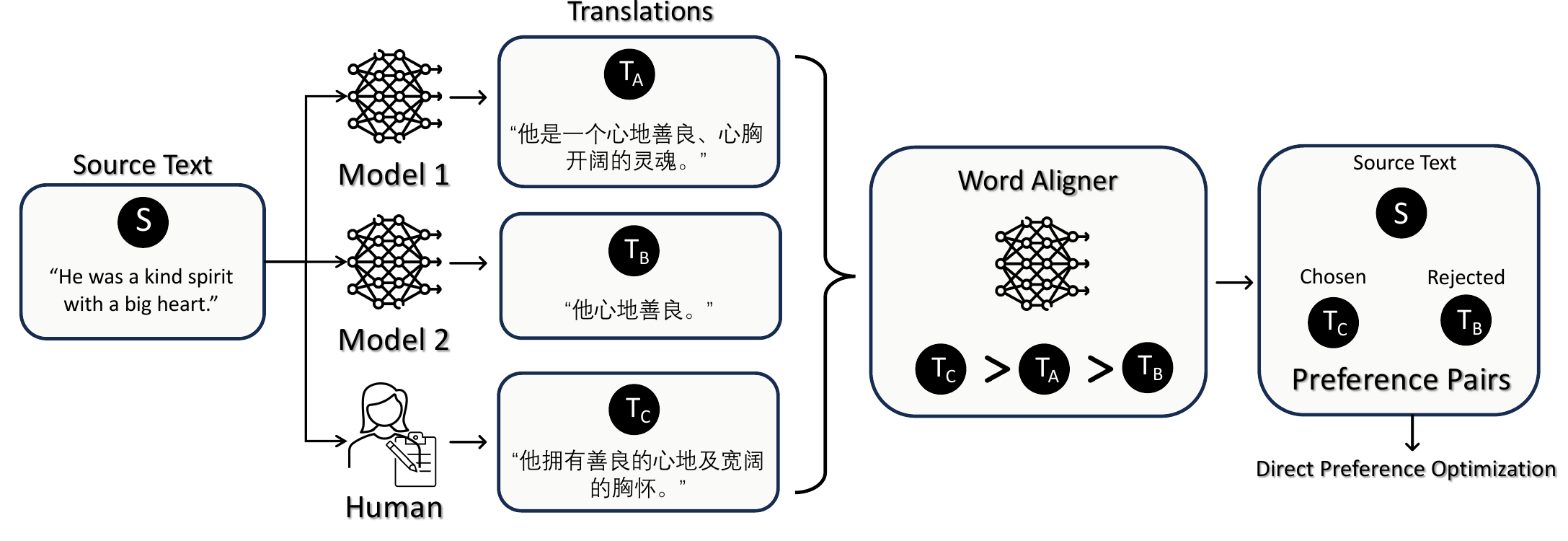}
    \caption{An illustration of \methodname{} framework. The source is first translated by multiple MT tools, including human translation. An external word aligner is then utilized to predict the coverage score for each translation. Finally, translation with the highest and lowest coverage score are selected as preference pairs for preference optimization.}
    \label{fig:framework}
\end{figure*}
\section{Proposed approach}
\label{sec:approach}

\subsection{Gathering translation candidates}
To steer the MT model to avoid \hao{} using preference optimization, we first need comparable but different translations. Starting with a source text $x$, we utilize $K$ methods to produce translations, notated as $\pi^1, ..., \pi^K$. Then we can get a set of translations $Y$, in which $y^k \in Y$ is obtained by $y^k = \pi^k(x)$ and $|Y|=K$.

\paragraph{Details of gathered translations} We start with the parallel training data in ALMA~\citep{alma}. This parallel data encompasses five language pairs with human translations in both directions: $cs \leftrightarrow en$, $de \leftrightarrow en$, $is \leftrightarrow en$, $zh \leftrightarrow en$ and $ru \leftrightarrow en$. We employ ISO 639 language codes\footnote{\url{https://en.wikipedia.org/wiki/List_of_ISO_639_language_codes}} to denote languages. Specifically, ``$cs$'' corresponds to Czech, ``$de$'' to German, ``$is$'' to Icelandic, ``$zh$'' to Chinese and ``$ru$'' and ``$en$'' to Russian and English, respectively. To generate the translations we require, this dataset is translated in both directions using two well-known MT tools, including DeepL\footnote{\url{https://www.deepl.com/en/translator}} and ChatGPT (\texttt{gpt-3.5-turbo-0613})\footnote{\url{https://platform.openai.com/docs/models/gpt-3-5-turbo}}.
The prompt we use to translate sentences is shown in Figure~\ref{fig:chatgpt_translation_prompt}.
The original human-written translation in the training set is also utilized. In particular, Icelandic ($is$) is not supported by DeepL, therefore, we use Google Translate\footnote{\url{https://cloud.google.com/translate/docs/basic/translate-text-basic}} as an alternative.

\input{figures/chatgpt_translation_prompt}
\subsection{Selecting chosen and rejected translation}
\label{sec:preferdata}
After obtaining the translation candidates $(y^1, ..., y^K)$, we use a state-of-the-art public word aligner, namely WSPAlign\footnote{\url{https://github.com/qiyuw/WSPAlign}}, to automatically annotate the degree of coverage for each translation. We follow the usage setting in the original paper of WSPAlign~\citep{wu-etal-2023-wspalign}. In particular, WSPAlign performs a bidirectional alignment and uses a threshold to filter out low-confident alignment of word pairs. Then, the ratio of the source words, \emph{that are aligned with at least one word}, in the translation is taken as the coverage score, which will be used for the following preference annotation.
The whole process to predict the coverage score is notated as $\mathrm{C}(\cdot,\cdot)$. Formally, the coverage score for a translation $y^k$ can be calculated by $\mathrm{C}(x,y^k) \in [0.0, 100.0]$. Subsequently, the preferred translation and the rejected translation are selected as follows:

\begin{gather}
\begin{aligned}
    y^w &= \mathop{\arg \max}\limits_{y^k \in Y}{\mathrm{C}(x,y^k)} \\
    y^l &= \mathop{\arg \min}\limits_{y^k \in Y}{\mathrm{C}(x,y^k)}
\end{aligned}
\end{gather}

where $y^w$ is the chosen translation and $y^l$ is the rejected one. Then a triplet $(x, y^w, y^l)$ is constructed for the following preference optimization.

\subsection{Filtering}
Note that the whole process of constructing the preference data is automatic, and the existing MT and word alignment models are not perfect. Even for human-annotated translation, quality is also an issue that cannot be ignored~\citep{cpo}, and can affect the performance of the model trained on it. Hence, noises are inevitable in both the translated texts and the preference choices. On the other hand, the MT tools we choose generally have good performance, it could happen that the generated translations are not diverse enough, leading to the preference signal being disrupted.
To improve the quality of the constructed preference datasets as much as possible, multiple strategies are applied to filter out potential bad training instances:
\begin{itemize}
    \item Remove the instance when the chosen and rejected translations only have a marginal difference in coverage score. The difference threshold is empirically set as 5.0, that is, $(x, y^w, y^l)$ is excluded from the dataset if $\mathrm{C}(x,y^w) - \mathrm{C}(x,y^l) < 5.0$.
    \item Remove the instance where the chosen and rejected translations are too semantically similar. Sentence embedding is a widely used technique for sentence similarity with low computation cost~\cite{gao-etal-2021-simcse, wu-etal-2022-pcl, xie2022stable, zhao-etal-2024-leveraging}.
    In particular, \texttt{LaBSE}~\cite{feng-etal-2022-language}\footnote{\url{https://huggingface.co/sentence-transformers/LaBSE}} is used in our experiments. We notate it as $\mathrm{LB}(\cdot)$. The similarity threshold is empirically set as 0.9, i.e. $(x, y^w, y^l)$ is excluded from the dataset if $\mathrm{sim}(\mathrm{LB}(y^w), \mathrm{LB}(y^w)) > 0.9$. $\mathrm{sim}(\cdot, \cdot) \in [0.0, 1.0]$ is cosine similarity.
    \item One possible failure case for word alignment is when the MT models directly copy the original texts, which is bad translation, but gets a high alignment score because the wrong translation is partially the same with the original texts. To remove this part of the noise, we calculate the BLEU score~\cite{papineni-etal-2002-bleu}\footnote{\url{https://github.com/mjpost/sacrebleu}} for the chosen translation and exclude it if the BLEU score $ > 20.0$.
\end{itemize}

\section{Details of dataset}
\label{sec:details_dataset}
\begin{figure}[ht]
    \centering
    \includegraphics[width=\linewidth]{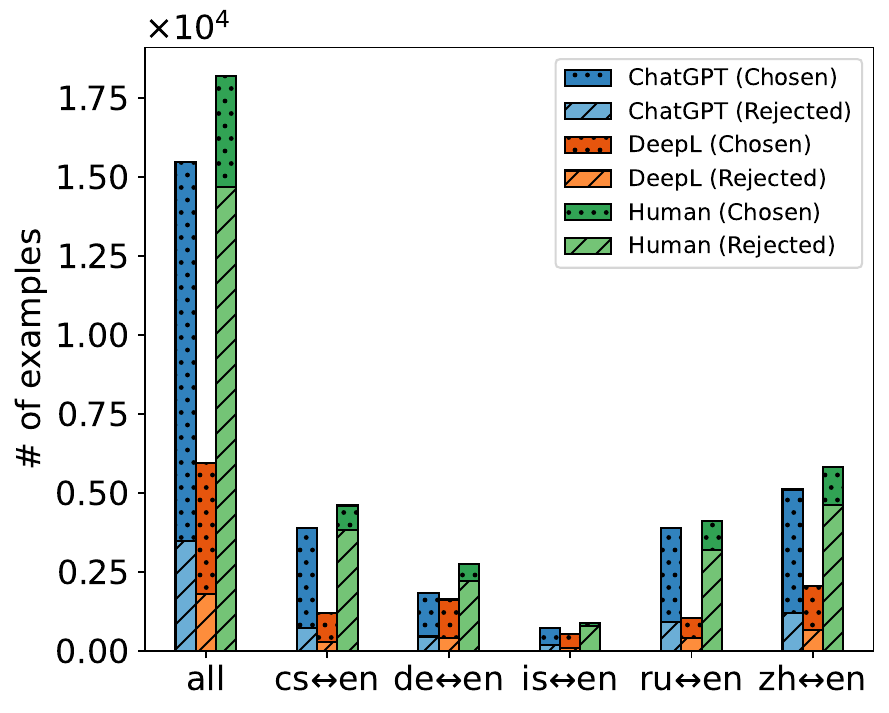}
    \caption{This figure illustrates the proportions of ``chosen'' and ``rejected'' preference pairs derived from three sources: ChatGPT, DeepL and Human. ``all'' represents the overall proportion for the aggregated dataset. $xx \leftrightarrow en$ is the subset pair of English and another language. Particularly, Google Translate is used for $is \leftrightarrow en$ as an alternative to DeepL.}
    \label{fig:chosen_vs_rejected}
\end{figure}

Figure~\ref{fig:chosen_vs_rejected} presents the varying proportions of the ``chosen'' and ``rejected'' preference pairs from three sources: ChatGPT, DeepL, and Human. The figure indicates that most of the ``chosen'' translations originate from ChatGPT, while a significant portion of human-written translations are ``rejected''. This observation supports the conclusion that human-written translations can also exhibit quality issues, as discussed in ALMA-R~\citep{cpo}. Examples in our constructed preference dataset are presented in \S\ref{sec:data_example}.

\subsection{Optimization LLM-based MT model}
The final step is to optimize the LLM-based MT model on our preference data. Direct preference optimization (DPO) \citep{dpo} is a simple but effective approach that directly optimizes the preference model on a pre-constructed static dataset. DPO has recently been applied to optimize LLM in preference data~\citep{tunstall2023zephyr,cpo} recently. We also utilize DPO as an optimization approach. Formally, the training objective is as follows,
\begin{equation}
    l = -\log\sigma(\beta\log\frac{\pi(y^w|x)}{\pi_{ref}(y^w|x)} - \beta\log\frac{\pi(y^l|x)}{\pi_{ref}(y^l|x)})
\end{equation}

where $\sigma$ is the sigmoid function, $\pi$ is the model to be optimized, and $\pi_{ref}$ is the reference model. We use \texttt{ALMA-13B}\footnote{\url{https://github.com/fe1ixxu/ALMA}} as our base model, i.e., the starting point of $\pi$, in the experiments. \texttt{ALMA-13B} is also used as a reference model $\pi_{ref}$, but note that $\pi_{ref}$ will not be updated during training.

%% file: figures/chatgpt_translation_prompt.tex
\begin{figure}[ht]
    \centering
    \includegraphics[width=0.5\textwidth]{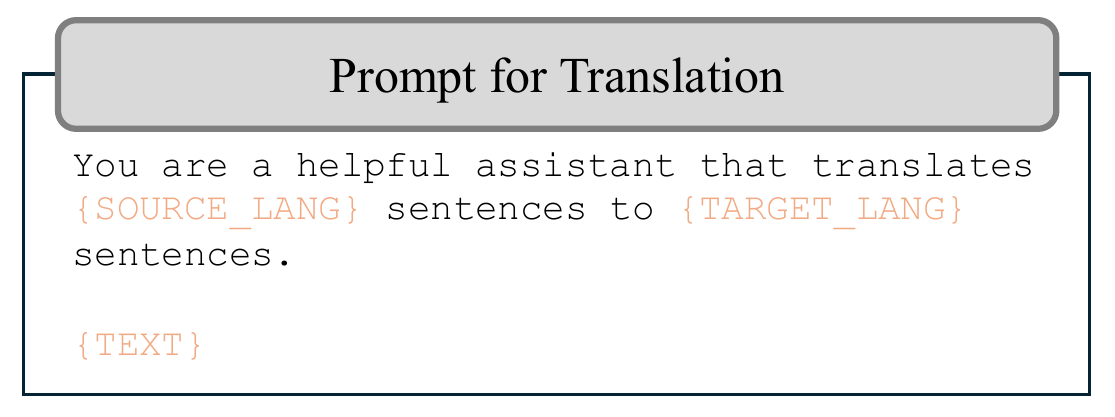}
    \caption{The prompt for translating sentences.}
    \label{fig:chatgpt_translation_prompt}
\end{figure}

%% file: sections/exp.tex
\input{figures/grid_figure}
\section{Evaluation}
\label{sec:eval}
\subsection{Baselines and evaluation datasets}
\label{sec:datasets}
We choose \texttt{ALMA-13B}\footnote{\url{https://huggingface.co/haoranxu/ALMA-13B}} as the baseline for all experiments in this paper, as well as the starting point of optimization.  ALMA~\citep{alma} was trained from Llama~\cite{touvron2023llama} in two steps: initial fine-tuning on monolingual data and subsequent fine-tuning on a small set of high-quality parallel data.
For fairly studying the effect of word alignment preference, we use the data used in the supervised fine-tuning in ALMA as the source dataset to construct our preference data in \S\ref{sec:approach}.
Specifically, the source data was collected from WMT'17 \citep{bojar-etal-2017-findings} to WMT'20 \citep{barrault-etal-2020-findings}, in addition to the development and text dataset from Flores-200 \citep{nllb2022}. After filtering, we finally make 20,074 and 2,226 preference triplets for training and development, respectively.
For evaluation, the test set is from WMT22, except that $is \leftrightarrow en$ is from WMT21. The remaining data from WMT21 (except $is \leftrightarrow en$) is used as the development set. Specifically, 3485, 4021, 2000, 3912, 4053 examples are included in the test set for $cs \leftrightarrow en$, $de \leftrightarrow en$, $is \leftrightarrow en$, $zh \leftrightarrow en$, and $ru \leftrightarrow en$, respectively.
The detailed experimental setup is introduced in \S\ref{sec:exp_setup}.

\paragraph{HalOmi}
In particular, we want to validate whether our proposed method is capable of mitigating \hao{} in MT. Hence, we also use HalOmi~\citep{dale-etal-2023-halomi} in the experiments. HalOmi is an evaluation benchmark for the detection of hallucination and omission in MT. It contains fine-grained sentence-level and token-level annotations of full and partial hallucinations and omissions that cover 18 language directions. Each instance in the data set was annotated in ``No \hao'', ``Small \hao'', ``Partial \hao'' or ``Full \hao'' by humans. In this paper, we use it to test the performance of GPT-4 as an evaluator. Details are in \S\ref{sec:eval_design}.

\subsection{The design of evaluation}
\label{sec:eval_design}
We focus on optimizing LLM-based MT models to avoid \hao{}. However, to our best knowledge, there is no benchmark measuring MT models specifically for this issue, making the evaluation very challenging. Improving the BLEU or COMET score does not necessarily mean reducing \hao{} because there are other factors such as mistranslation and fluency. In addition, hallucination is relatively infrequent, although very severe once it does occur. To intuitively validate whether our approach is capable of mitigating \hao{} in MT, we design several evaluation strategies in this section.

\paragraph{Select hard instances.}
We first select instances that the baseline model does not perform well on. This subset of instances is labeled as \textit{hard instances} in this work. The subset of the remaining examples is labeled as \textit{easy instances}. Specifically, $N$ instances with the lowest COMET score are selected from the test set for each translation direction. As hard examples tend to include more \hao{}, we report the comparison of models on hard examples and remaining examples, respectively.
In the experiment, we sample three subsets where $N=100$, $N=200$ and $N=500$. The experimental analysis can be found in \S\ref{sec:eval_hard}.
Note that the hard instances are only selected for evaluation. We do not differentiate hard or easy instances in the training set. Only word alignment signal is used to select preferred dataset for a fair comparison.

\input{figures/prompt}
\input{tables/gpt_ho}
\paragraph{Utilize LLM as the evaluator for \hao{}.}
Besides the BLEU and COMET in hard instances, a direct estimate of the degree of \hao{} in translation is still needed. As we mentioned earlier that improving the BLEU and COMET score does not necessarily mean reducing \hao{} because there are other factors such as mistranslation and fluency, we utilize the generalization and reasoning ability of LLM~\citep{Kojima2022LargeLM, Mitchell2023DetectGPTZM, Wei2023ZeroShotIE} to achieve this direct evaluation.
We use one of the most powerful LLM, \texttt{gpt-4-0613}\footnote{\url{https://platform.openai.com/docs/models/gpt-4-turbo-and-gpt-4}}, as an evaluator. LLM is prompted to check whether the given translation has \hoo{} referring to the given source texts. A coverage score between 0 and 100 is output as the degree metric. The prompt used is shown in Figure~\ref{fig:prompt}.

\paragraph{Is LLM really capable of evaluating \hao{} in MT?}
Despite the fact that LLMs have shown impressive zero-shot performance in various tasks \citep{Kojima2022LargeLM, Mitchell2023DetectGPTZM, Wei2023ZeroShotIE}, the assessment of LLM in the evaluation of \hao{} is still important because it has not been widely used on this task.
We use HalOmi datasets introduced in \S\ref{sec:datasets} to assess this ability of GPT-4.
The examples in $de \leftrightarrow en$, $zh \leftrightarrow en$, and $ru \leftrightarrow en$ are selected, then GPT-4 is used to predict the coverage score for these examples.

Table \ref{tab:gpt_ho} shows the average score of the degree of coverage predicted by GPT-4. The examples from HalOmi are divided into three subsets according to the labels. We merged the ``Partial \hao'' and ``Small \hao'' in the original because the number of examples in these two categories is small. It clearly demonstrates that examples annotated as ``No \hao'' have a higher coverage score predicted by GPT-4 and those in ``Full \hao'' have an extremely low coverage score. As a result, using GPT-4 is an effective way to assess whether a translation has the problem of \hoo.

\input{tables/gpt_eval_results}
\input{tables/human_eval}
\section{Experimental results}
\label{sec:exp}
\subsection{Evaluation on hard instances}
\label{sec:eval_hard}
In \S\ref{sec:eval_design} we introduce how to select hard instances from the test set and explain why hard instances are suitable to assess \hao. In this section, we evaluate our model on these hard instances and the remaining examples, respectively. Figure~\ref{fig:main_results} demonstrates the results when the number of hard instances $N=100, 200$, and $500$, respectively.
The following findings can be concluded:
\begin{itemize}
    \item \methodname{} consistently outperforms the baseline in hard instances in most translation directions, for both BLEU and COMET metrics.
    \item \methodname{} generally reaches comparable performance compared to baseline for both BLEU and COMET.
    \item With increasing the number of hard instances, the improvement gained by \methodname{} decreases.
\end{itemize}
These results indicate that \methodname{} mitigates \hao{} to a certain extent, because these issues are more likely to occur in hard instances. In addition, our model also remains competitive to the baseline in the remaining easy instances. It is reasonable that there is no significant difference because the compared models are generally good. The challenging part should be in the hard ones.
Moreover, it is observed that with increasing $N$, the improvement gets narrower. The reason is that more relatively easy instances are included in the subset. This is another evidence that \methodname{} provides gains particularly for \hao{} in MT. The specific numeric results and the overall results for the entire test set are shown in \S\ref{sec:all_results}.

\subsection{Direct evaluation of \hao{} by GPT-4}
\label{sec:eval_llm}
In addition to improving BLEU and COMET in hard examples prone to \hao, direct evaluations are also necessary to confirm the effectiveness of \methodname{}.
In \S\ref{sec:eval_design} we have verified the usefulness of GPT-4 as an evaluator with experiments. In this section, we prompt GPT-4 to directly predict a coverage score as a metric for \hao{}. The results are demonstrated in Table~\ref{tab:gpt_eval_result}. The reported number is the average of the coverage scores in hard examples.
The results show that \methodname{} outperforms the baseline in all directions except $en \leftrightarrow is$. In the overall average score across all translation directions, \methodname{} outperforms the baseline model by \textbf{4.96}, \textbf{1.63} and \textbf{1.24} when N=100, 200 and 500, respectively. The trend is similar to that in \S\ref{sec:eval_hard}, directly indicating that the LLM-based MT model avoids \hao{} with the word-aligned preference.

\subsection{Human evaluation}
Although the validity of GPT-4 as evaluator for \hao{} has been demonstrated in \S\ref{sec:eval_design} and Table~\ref{tab:gpt_ho}, we conduct a human evaluation to further verify our findings, as LLM could still be unreliable. The subset of ``N=100'' on ``zh-en'' is selected. Three volunteers who speak Chinese and English are asked to assess the quality of the translation and the degree of \hao{} for the baseline and our model, without knowing which model generates the translations. Table~\ref{tab:human_eval} demonstrates the results. In general, our model generates better translation in 39.66\% of the examples, while the percentage for ALMA is 11.33\%. Furthermore, it is observed that with DPO on word-alignment preferred data fine-tuning, the degree of both \hao{} decreases. Specifically, the percentage of ``no hallucination'' increases from 64\% to 75.66\%, and that of ``small, partial, and full hallucination'' decreases accordingly. The decrease in omission is more distinct, in which the percentage of ``no omission'' increase by 24\%. Notably, for both hallucination and omission, the percentage of ``full \hao{}'' has decreased to 0 for our model. These results indicate that omission is more frequent than hallucination, and \methodname{} can mitigate them in the LLM-based MT model.

\subsection{Ablation study}
\label{sec:ablation}
\input{figures/ablation}
In this section, we conduct in-depth investigation for our word alignment preference, as we use the same training data as our baseline ALMA, i.e., human translation, but extra translations from DeepL and ChatGPT are included to conduct our preference data. To investigate where the improvement comes from, we introduce two variants without preference tuning to compare with \methodname{}. 
\begin{itemize}
    \item \textit{FT\_reject}: directly fine-tuning ALMA with the rejected translations in the dataset.
    \item \textit{FT\_prefer}: directly fine-tuning ALMA with the preferred translations in the dataset.
\end{itemize}
The comparison is demonstrated in Figure~\ref{fig:ablation}.

\paragraph{Does the preferred data truly contribute more to training?}
It is observed that \textit{FT\_prefer} significantly outperforms \textit{FT\_reject} in both hard and easy instances. This suggests that \methodname{} effectively selects samples, improving translation quality. It highlights the importance of selecting high-quality training data, as even human-annotated data can be flawed~\cite{alma}.

\paragraph{Is DPO preference tuning necessary?}
The filled area highlights the necessity of preference tuning with DPO. While \textit{FT\_prefer} performs competitively in hard instances, it significantly underperforms \methodname{} and ALMA in easy instances, limiting its practicality. The possible reason for the different performance in the hard and easy instances can be the direct fine-tuning, which focusing solely on preferred data without comparing it to rejected examples can lead to overfitting to word-aligned preferences, neglecting overall translation quality.

%% file: figures/grid_figure.tex
\begin{figure*}[ht]
\setlength\tabcolsep{2pt}
\centering
\scriptsize
\resizebox{\textwidth}{!}{
\begin{tabular}{@{} c M{.23\textwidth} M{.23\textwidth} M{.23\textwidth} M{.23\textwidth} @{}}
 ~ & BLEU / Hard & BLEU / Easy & COMET / Hard & COMET / Easy \\ \hline
N=100 &
 \includegraphics[width=.23\textwidth]{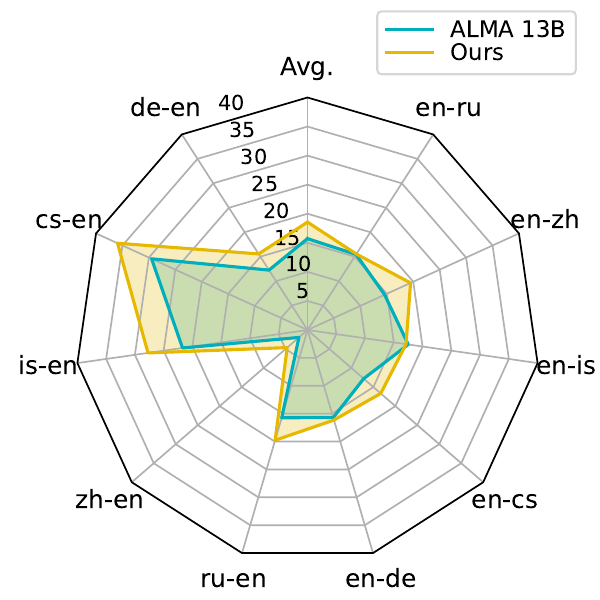} &
 \includegraphics[width=.23\textwidth]{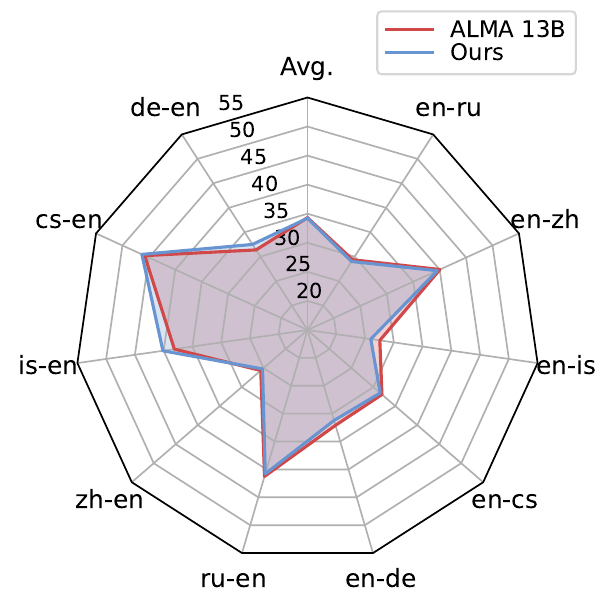} &
 \includegraphics[width=.23\textwidth]{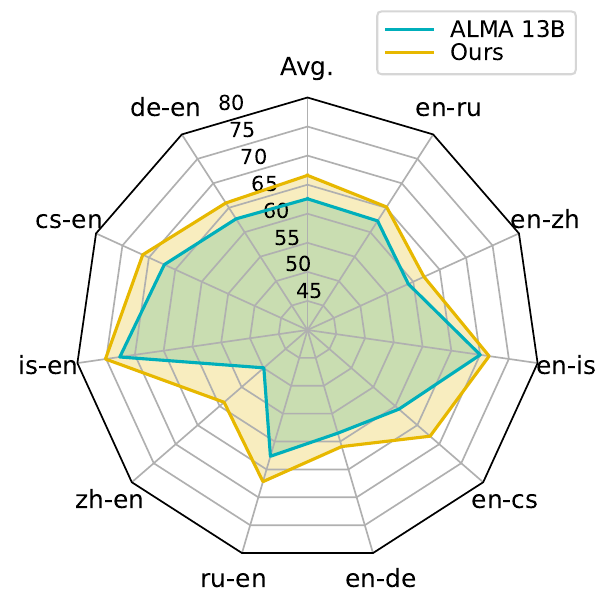} &
 \includegraphics[width=.23\textwidth]{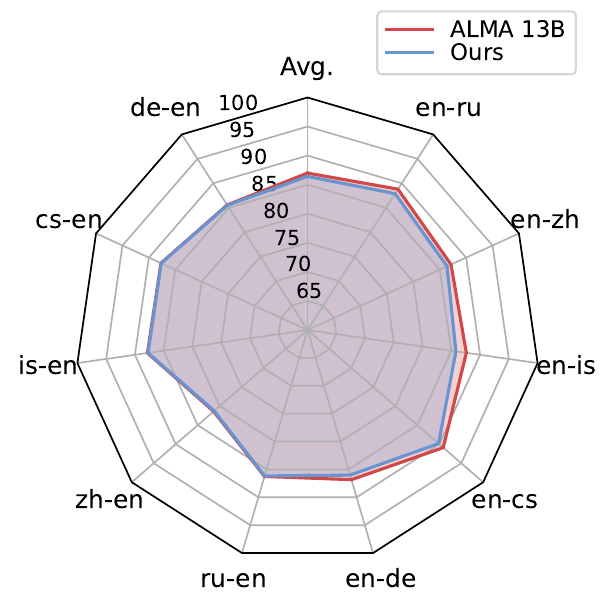} \\
N=200 &
\includegraphics[width=.23\textwidth]{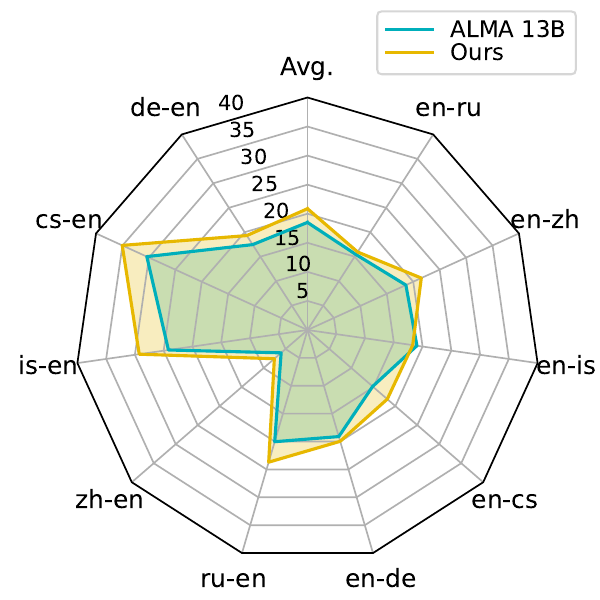} &
 \includegraphics[width=.23\textwidth]{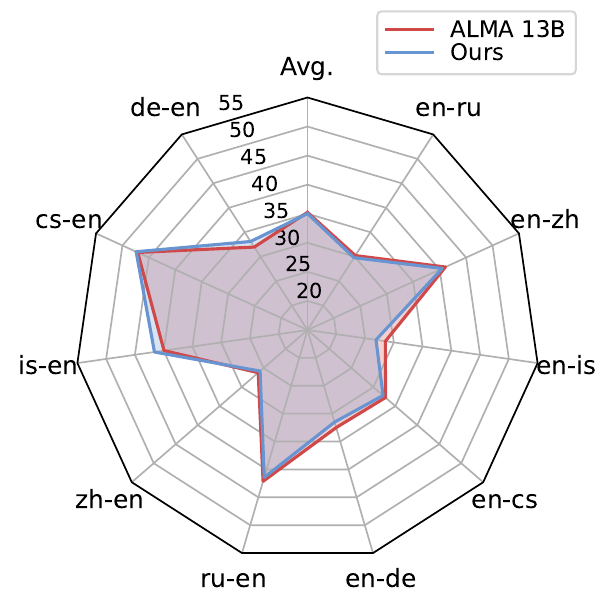} &
 \includegraphics[width=.23\textwidth]{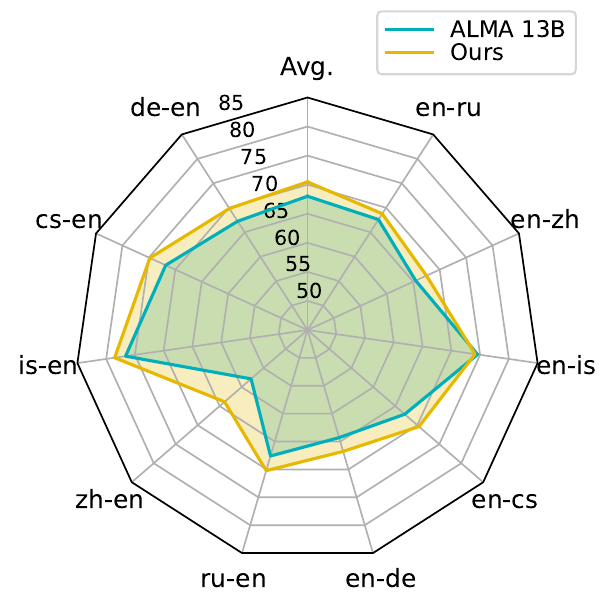} &
 \includegraphics[width=.23\textwidth]{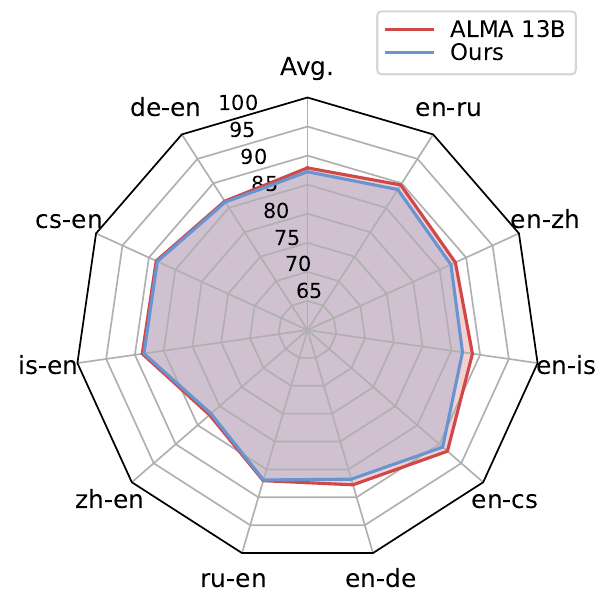} \\
N=500 &
\includegraphics[width=.23\textwidth]{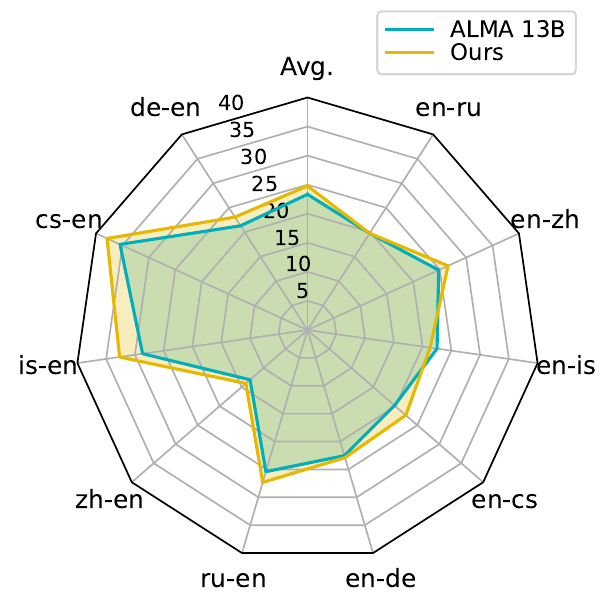} &
 \includegraphics[width=.23\textwidth]{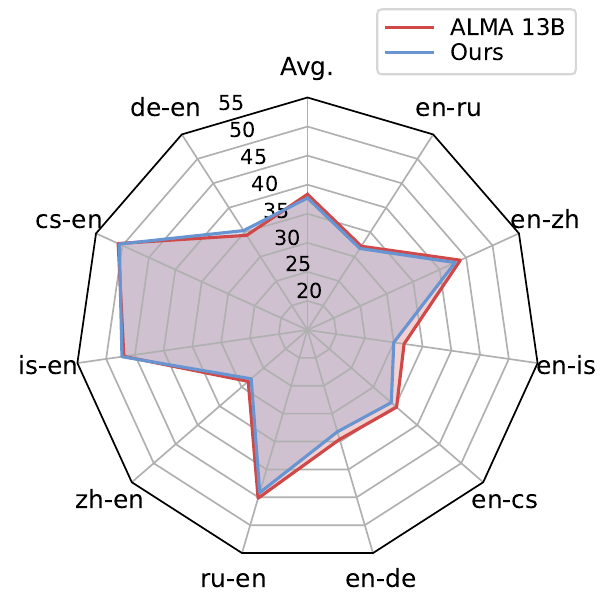} &
 \includegraphics[width=.23\textwidth]{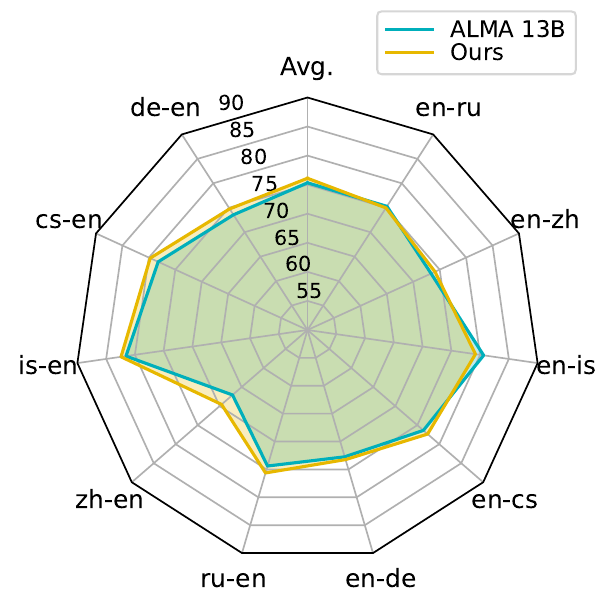} &
 \includegraphics[width=.23\textwidth]{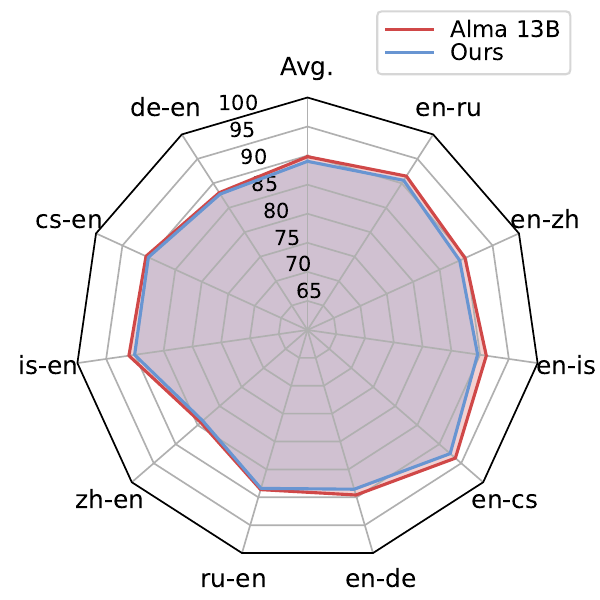} \\
\end{tabular}
}
\caption{Comparison of \methodname{} and baseline in hard and easy instances. $N$ instances with the lowest COMET score by the baseline are selected from the test set as hard instances, and the remaining are easy instances. Results when $N=100$, $200$ and $500$ are presented. Refer to \S\ref{sec:all_results} for the full numeric results of the entire test.}
\label{fig:main_results}
\end{figure*}

%% file: figures/prompt.tex
\begin{figure}[ht]
    \centering
    \includegraphics[width=\columnwidth]{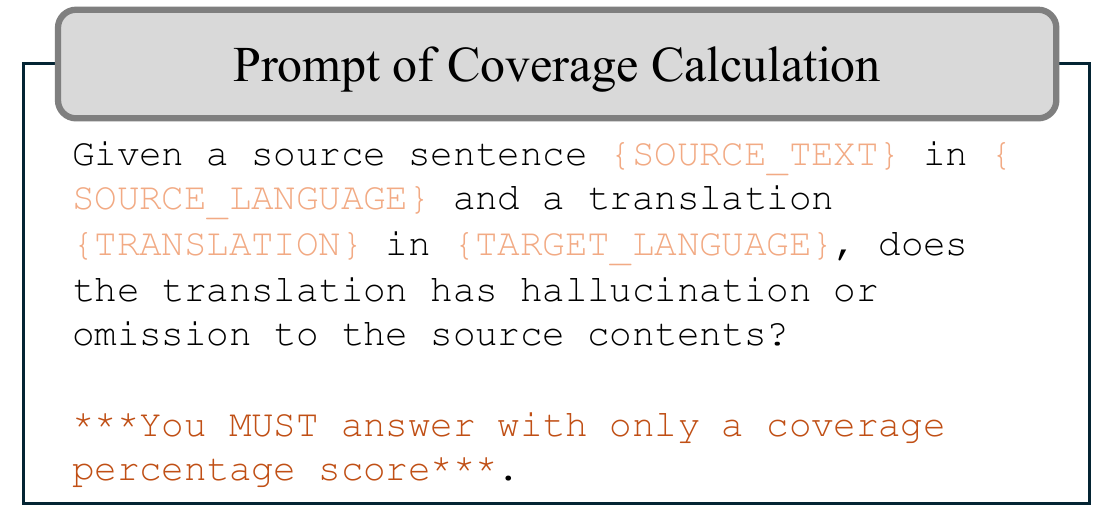}
    \caption{Prompt to calculate the coverage score.}
    \label{fig:prompt}
\end{figure}

%% file: tables/gpt_ho.tex
\begin{table*}[ht]
\centering
\small
\begin{tabularx}{\textwidth}{X|Y|Y|Y|Y|Y|Y}
\hline
\multirow{2}{*}{}   & \multicolumn{3}{c|}{Hallucination} & \multicolumn{3}{c}{Omission} \\ \cline{2-7} 
                    & No        & Partial     & Full     & No      & Partial   & Full   \\ \hline
\# of examples      & 817       & 42          & 65       & 627     & 237       & 60     \\ \hline
Avg.  score & 84.19     & 45.95       & 3.84     & 87.97   & 66.28     & 1.66   \\ \hline
Pearson Corr.       & \multicolumn{3}{c|}{0.5969}        & \multicolumn{3}{c}{0.5686}   \\ \hline
\end{tabularx}
\caption{Average coverage score calculated by GPT-4 for different level of \hoo{}. The Pearson Correlation between the annotated labels and GPT-4 coverage scores is also reported. Ideally, higher score should correlate to less \hao{}.}
\label{tab:gpt_ho}
\end{table*}

%% file: tables/gpt_eval_results.tex
\begin{table*}[ht]
\centering
\resizebox{\textwidth}{!}{
\begin{tabular}{cccccccccccc}
\hline
\multicolumn{1}{c|}{}         & de-en          & cs-en          & is-en          & zh-en          & ru-en          & en-de          & en-cs          & en-is          & en-zh          & \multicolumn{1}{l|}{en-ru}          & Avg.           \\ \hline
\multicolumn{12}{c}{N=100}                                                                                                                                                                                                                    \\ \hline
\multicolumn{1}{c|}{Baseline} & 94.30          & 92.95          & 94.90          & 63.08          & 89.85          & 92.85          & 82.75          & \textbf{97.05} & 84.65          & \multicolumn{1}{l|}{90.53}          & 88.29          \\
\multicolumn{1}{c|}{\quad +\methodname{}}     & \textbf{95.85} & \textbf{94.65} & \textbf{96.05} & \textbf{80.23} & \textbf{91.75} & \textbf{96.25} & \textbf{91.85} & 96.10          & \textbf{92.90} & \multicolumn{1}{l|}{\textbf{96.87}} & \textbf{93.25\small{(+4.96)}} \\ \hline
\multicolumn{12}{c}{N=200}                                                                                                                                                                                                                    \\ \hline
\multicolumn{1}{c|}{Baseline} & 95.71          & 95.05          & 95.45          & 74.83          & 92.83          & 94.20          & 89.95          & \textbf{97.70} & 89.19          & \multicolumn{1}{l|}{94.25}          & 91.92          \\
\multicolumn{1}{c|}{\quad +\methodname{}}     & \textbf{97.10} & \textbf{96.55} & \textbf{97.48} & \textbf{85.63} & \textbf{95.53} & \textbf{95.18} & \textbf{91.84} & 96.73          & \textbf{92.81} & \multicolumn{1}{l|}{\textbf{96.66}} & \textbf{94.55\small{(+2.63)}} \\ \hline
\multicolumn{12}{c}{N=500}                                                                                                                                                                                                                    \\ \hline
\multicolumn{1}{c|}{Baseline} & 97.18          & 96.74          & 97.29          & 87.85          & 96.16          & 97.35          & 94.46          & 98.21          & 91.64          & \multicolumn{1}{l|}{96.10}          & 95.30          \\
\multicolumn{1}{c|}{\quad +\methodname{}}     & \textbf{98.10} & \textbf{97.79} & \textbf{98.12} & \textbf{90.76} & \textbf{97.82} & \textbf{97.36} & \textbf{96.05} & \textbf{98.22} & \textbf{94.07} & \multicolumn{1}{l|}{\textbf{97.13}} & \textbf{96.54\small{(+1.24)}} \\ \hline  
\end{tabular}
}
\caption{Coverage score output by GPT-4. The range of the score is $[0.0,100.0]$. The average score is reported for each translation direction. Higher scores are highlighted in bold.}
\label{tab:gpt_eval_result}
\end{table*}

%% file: tables/human_eval.tex
\begin{table*}[ht]
\centering
\resizebox{\textwidth}{!}{
\centering
\begin{tabular}{c||c||cccc||cccc}
\hline
     & \multirow{2}{*}{Translation Quality} & \multicolumn{4}{c||}{Hallucination}   & \multicolumn{4}{c}{Omission}         \\ \cline{3-10} 
     &                                      & No      & Small   & Partial & Full   & No      & Small   & Partial & Full   \\ \hline
Baseline & 11.33\%                              & 64.00\% & 21.00\% & 11.33\% & 3.66\% & 56.00\% & 25.33\% & 13.66\% & 4.33\% \\
\quad +\methodname{} & \textbf{39.66\%}                              & \textbf{75.66\%} & \textbf{17.33\%} & \textbf{7.00\%}  & \textbf{0.00\%} & \textbf{80.00\%} & \textbf{16.66\%} & \textbf{5.33\%}  & \textbf{0.00\%} \\ \hline
\end{tabular}
}
\caption{Human evaluation on ``zh-en'' when N=100. Translation quality is the measured by ratio of examples where \methodname{} beats the baseline. The remaining columns present the ratio of examples in which the corresponding degree of \hoo{} occurs. Better model is highlighted with bold fonts.}
\label{tab:human_eval}
\end{table*}

%% file: figures/ablation.tex
\begin{figure*}[ht]
    \centering
    \begin{subfigure}{\columnwidth}
        \centering
        \includegraphics[width=\textwidth]{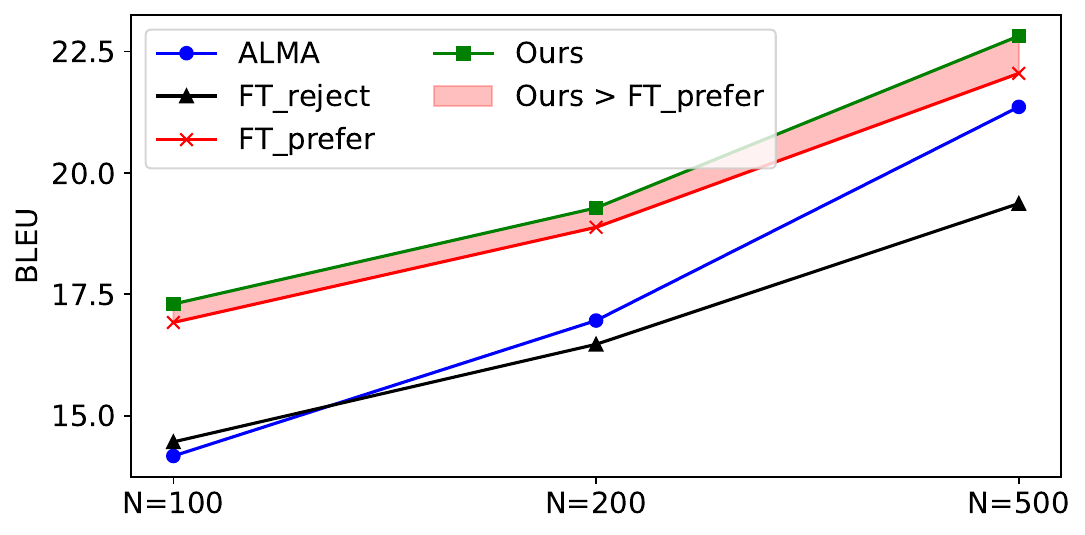}
        \caption{Hard instances}
    \end{subfigure}
    \begin{subfigure}{\columnwidth}
        \centering
        \includegraphics[width=\textwidth]{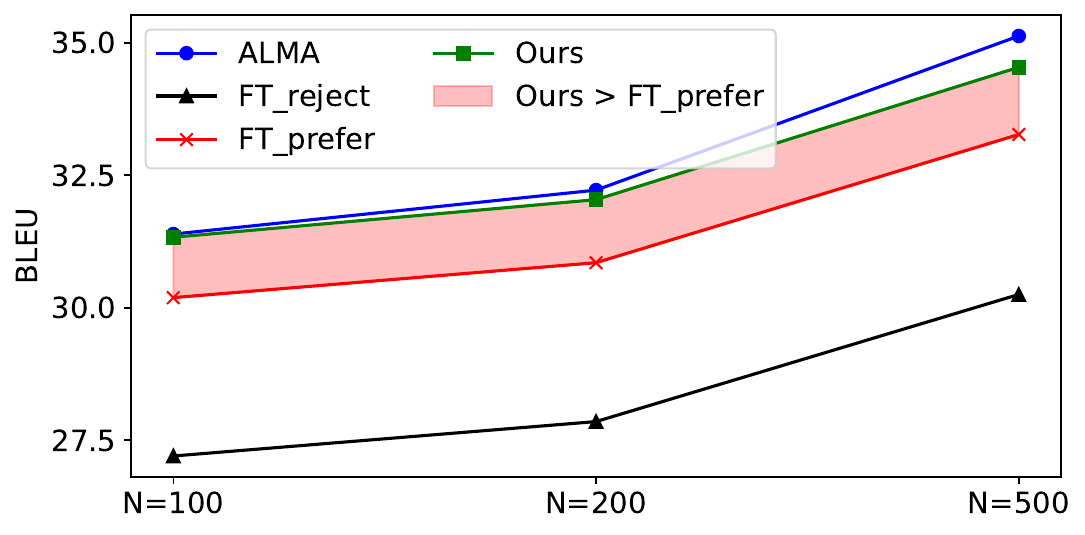}
        \caption{Easy instances}
    \end{subfigure}
    \begin{subfigure}{\columnwidth}
        \includegraphics[width=\textwidth]{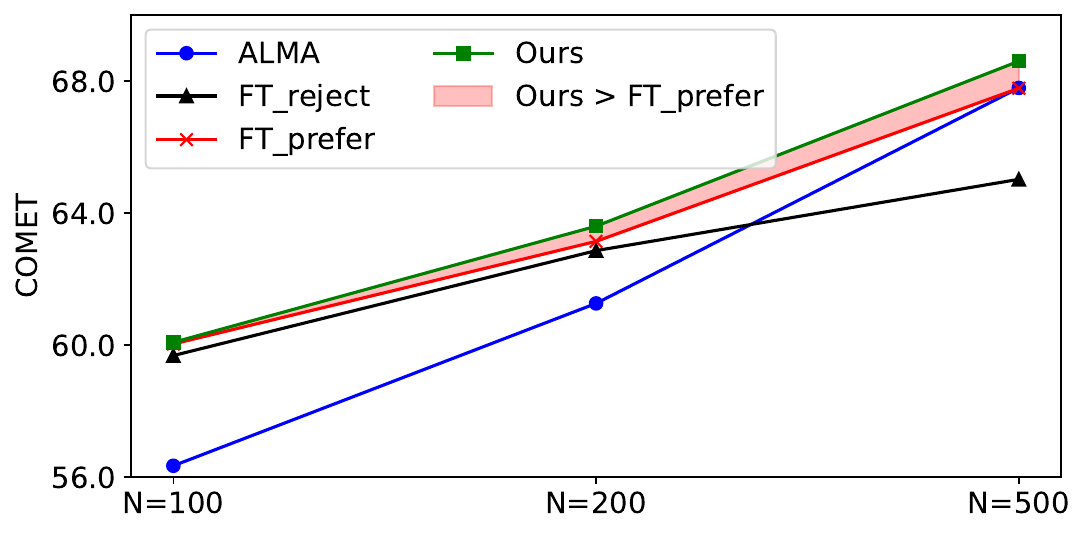}
        \caption{Hard instances}
    \end{subfigure}
    \begin{subfigure}{\columnwidth}
        \includegraphics[width=\textwidth]{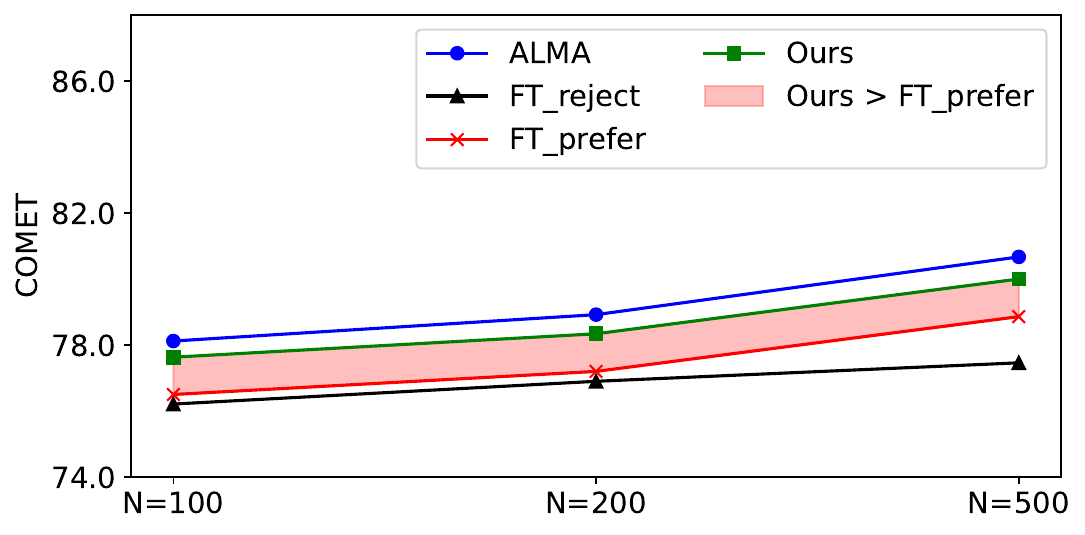}
        \caption{Easy instances}
    \end{subfigure}
    \caption{Ablation study. Results in BLEU is demonstrated. Higher BLEU is better. For fair comparison the range of y-axis are the same for hard instances and easy instances.}
    \label{fig:ablation}
\end{figure*}

%% file: sections/appendix.tex
\section{Experimental setup}
\label{sec:exp_setup}
The implementation from alignment-handbook\footnote{\url{https://github.com/huggingface/alignment-handbook}} is used for the training of DPO. The learning rate is searched based on performance on development set and set to 5e-6. LoRA~\citep{Hu2021LoRALA} is used. $r$ is set as 16 and $\beta$ is set as 0.1. We train the model for 1 epoch and fix the random seed to 42. The model is trained on 4 $\times$ Nvidia A100 80G and the total batch size is 64. For evaluation, we use the implementation of ALMA\footnote{\url{https://github.com/fe1ixxu/ALMA}} to calculate the BLEU and COMET scores.

\input{tables/data_example}

\section{Example analysis}
\subsection{Examples of the preference dataset}
\label{sec:data_example}
Table~\ref{tab:data_example} includes three examples in our dataset, in which the source sentence, the chosen and rejected translations are shown. Refer to \S\ref{sec:details_dataset} for a detailed construction of the dataset. \textbf{Example 1}: the rejected translation is from human annotation, in which it repeats the term of ``I think'' unnaturally. The possible reason could be the resource of the parallel data, e.g., direct collection from transcriptions. \textbf{Example 2}: ``Fuller'' is omitted by human annotation while translated by DeepL. \textbf{Example 3}: the chosen translation is from gpt-3.5-turbo that completely translates the source sentence. In contrast, the translation by DeepL omits the first half.
\input{tables/trans_example}

\subsection{Translation examples}
\label{sec:trans_example}
Table~\ref{tab:trans_example} shows illustrative comparison between translations from the baseline and our model. \textbf{Example 1}: ``in HBO's 'The Gilded Age'" in the source sentence is omitted by the baseline. In contrast, our model successfully translate the corresponding part into Chinese. \textbf{Example 2}: the baseline generates ``\chinese{卡扣} (fastening)'' infinitely in translation. This type of hallucination also occurs in other LLM applications, which emphasizes the need to address the hallucination issue in LLM-based MT models. \textbf{Example 3}: ``\chinese{等到什么时候} (when to wait)'' is omitted by the baseline model while our model translate that into ``how long I have to wait'' properly. 

On the other hand, WAP could also fail in some cases. \textbf{Example 4}: Although the baseline omits ``pictures'' and ``box,'' which our model successfully translates, the translation of our model is not fully correct. The source is ``in the box with frame,'' but our model's translation is ``\chinese{画框在盒子里} (frame in the box).'' \textbf{Example 5}: Although our model translates ``pot (\chinese{锅})'' that is omitted by the baseline, the meaning of the sentence is incorrect. The source means "This pot is a good buy," but our translation is "This pot is worth buying."
In general, our model performs well in terms of coverage, which is more related to hallucination and omission; however, the translation quality does not necessarily improve accordingly. The study of preference signals for both overall translation quality and reducing hallucination and omission is worth exploring.


\section{Overall MT Performance}
\label{sec:all_results}
\input{tables/all_results}

Table~\ref{tab:all_results} shows the numeric results in Figure~\ref{fig:main_results}, in which boxes on a blue background highlight the cases where our model outperforms the baseline by a margin $>1.0$, and the boxes in red are the opposite. Boxes without background indicate the cases when our model and the baseline have competitive performance where the margin $<1.0$.

In addition to the main findings in \S\ref{sec:eval_hard} that our model generally performs better in harder instances, from the results it can also be observed that our model particularly performs worse on ``\textit{en-is}'' than in other translation directions. The reason could be that Icelandic is a low-resource language and we used external tools such as WSPAlign and Google Translate to build the training data. Hence, the relatively unreliable performance of external tools on low-resource languages can induce noises in our training data. This could be a future direction for building more reliable word alignment signals and particular research on low-resource languages.

Table~\ref{tab:all_results} reports the overall performance when we do not split the dataset into the hard and easy subset. The results show that our model and ALMA have generally competitive performance. Specifically, if we only consider the margin larger than 1.0, our model outperforms ALMA on \textit{de-en} and \textit{is-en} in BLEU while ALMA performs better on \textit{en-is} in both BLEU and COMET. In particular, a significance test is conducted to investigate numeric degradation when all instances are included. We utilize bootstrap sampling from example-wise COMET scores with 100,000 iterations and calculate the p-value. Based on the results of the significance test, there is no statistical significance when the margin is greater than 0.25, indicated by a p-value larger than 0.05.  
This suggests that our approach does not degrade the general performance by a margin of 0.25 or more, while improving that on hard instances by a large margin of 3.47. Note that the focus of this work is the problem of hallucination and omission, general metrics for MT are only partially related to our evaluation. The evaluation by LLM and humans is also important, as we discussed in \S\ref{sec:eval_design}.

%% file: tables/data_example.tex
\begin{table*}[ht]
\begin{tabularx}{\textwidth}{l|X|c}
\hline
\rowcolor{lightgray}\multicolumn{2}{l|}{\textbf{Example 1 (Chinese-English)}} & \textbf{Coverage Score}\\ \hline
source & \begin{CJK*}{UTF8}{gbsn}“我想，在考虑重播时，可以解决这个问题”，Coker 说道。\end{CJK*} & --\\ \hline
chosen (gpt-3.5) & "I think, when considering replay, this issue can be resolved," Coker said. & 94.03 \\ \hline
rejected (human) & "\redmark{I think that when I think about} the replay, \redmark{I think that} we can probably work it out," Coker said. & 79.87 \\
\hline
\rowcolor{lightgray}\multicolumn{2}{l|}{\textbf{Example 2 (Chinese-English)}} & \textbf{Coverage Score} \\ \hline
source & \begin{CJK*}{UTF8}{gbsn}\redmark{富勒}在政变图谋失败后\end{CJK*} & -- \\ \hline
chosen (deepl) & \bluemark{Fuller} after the failed coup attempt & 83.76 \\ \hline
rejected (human) & After the failure of the attempted coup,  & 59.59 \\ \hline
\rowcolor{lightgray}\multicolumn{2}{l|}{\textbf{Example 3 (English-Chinese)}} & \textbf{Coverage Score} \\ \hline
source & \redmark{Originally a one-bedroom property with a convoluted layout - you had to walk through the kitchen to get to the bedroom} - Joanne wanted to add storage space and a mezzanine to make the most of the generous ceiling height.' & -- \\ \hline
chosen (gpt-3.5) & \begin{CJK*}{UTF8}{gbsn} \bluemark{最初是一个一居室的房产，布局错综复杂 - 你必须穿过厨房才能到达卧室} - 然而乔安妮想要增加存储空间和一个夹层，以充分利用宽敞的天花板高度。 \end{CJK*} & 83.76 \\ \hline
rejected (deepl) & \begin{CJK*}{UTF8}{gbsn} 乔安妮希望增加储藏空间和一个夹层，充分利用宽敞的天花板高度。\end{CJK*} & 69.97 \\ \hline
\end{tabularx}
\caption{Examples in the preference dataset. The hallucination in rejected examples and omission in the source sentence are highlighted with \redmark{\textit{\quad}}. The corresponding contents that are omitted in the rejected example are highlighted with \bluemark{\textit{\quad}} in the chosen example. The coverage is calculated by word aligner, refer to \S\ref{sec:approach} for details.}
\label{tab:data_example}
\end{table*}

%% file: tables/trans_example.tex
\begin{table*}[ht]
\begin{tabularx}{\textwidth}{l|X|c}
\hline
\rowcolor{lightgray}\multicolumn{2}{c}{\textbf{Example 1 (English-Chinese})} & \textbf{Coverage Score}\\ \hline
Source & Sunday Best: Enter 1880s New York \redmark{in HBO's "The Gilded Age"} & -- \\ \hline
Translation (Baseline) & \chinese{周日最佳：进入 1880 年代的纽约} & 70.0\\ \hline
Translation (Ours) & \chinese{周日最佳：进入 1880 年代的纽约 \bluemark{，在 HBO 的《金碧辉煌时代》}} & 100.0 \\ \hline
\rowcolor{lightgray}\multicolumn{2}{c}{\textbf{Example 2 (English-Chinese)}} & \textbf{Coverage Score}\\ \hline
Source & Liner Fastening and Hanging Tabs Inner tabs are provided to keep a loose liner in position, corresponding in position with the tabs we provide on our liners. & -- \\ \hline
Translation (Baseline) & \chinese{粘贴和悬挂\redmark{卡扣的内部卡扣用于保持卡扣卡扣卡扣卡扣卡扣卡扣卡扣卡扣卡扣卡扣卡扣卡扣卡扣卡扣卡扣卡扣卡扣卡扣卡扣卡扣卡扣卡扣卡扣卡扣卡扣卡扣卡扣卡扣卡扣卡扣卡扣卡扣卡扣卡扣卡扣卡扣卡扣}} & 0.0 \\ \hline
Translation (Ours) & \chinese{内固定和悬挂标签内固定和悬挂标签用于保持薄膜在位，与我们提供的标签对应。} & 60.0 \\ \hline
\rowcolor{lightgray}\multicolumn{2}{c}{\textbf{Example 3 (Chinese-English)}} & \textbf{Coverage Score}\\ \hline
Source & \chinese{不知道要\redmark{等到什么时候}} & -- \\ \hline
Translation (Baseline) & I don't know when & 90.0 \\ \hline
Translation (Ours) & I don't know \bluemark{how long I have to wait} & 100.0 \\ \hline
\hline
\multicolumn{2}{c}{\textbf{Example 4 (English-Chinese)}} \\ \hline
Source & \redmark{Pictures come in} a sturdy carded envelope \redmark{in the box with the frame.} \\ \hline
Translation (Baseline) & \chinese{画框附带一个坚固的卡片盒。} \\ \hline
Translation (Ours) & \chinese{画框\bluemark{在盒子里，图片}放在一个坚固的卡片盒里。} \\ \hline
\hline
\multicolumn{2}{c}{\textbf{Example 5 (Chinese-English)}} \\ \hline
Source & \chinese{这个\redmark{锅}买得好值。} \\ \hline
Translation (Baseline) & It's a good buy. \\ \hline
Translation (Ours) & This \bluemark{pot} is worth buying. \\ \hline
\end{tabularx}
\caption{Translation Examples. The hallucination in translation by the baseline and the omission in the source sentence are highlighted with \redmark{\textit{\quad}}. The corresponding contents that are omitted from the baseline are highlighted with \bluemark{ } in our translation. The coverage is calculated by GPT-4, refer to \S\ref{sec:eval_design} for details.}
\label{tab:trans_example}
\end{table*}

%% file: tables/all_results.tex
\begin{table*}[ht]
\resizebox{\textwidth}{!}{
\centering
\begin{tabular}{cccccccccccc}
\hline
Model-\scriptsize{Metric} & de-en & cs-en & is-en & zh-en & ru-en & en-de & en-cs & en-is & en-zh & en-ru & Avg.\\
\hline
\multicolumn{11}{c}{\textit{N=100}} \\
\cline{2-11}
\multicolumn{11}{c}{\textit{Easy instances}} \\
ALMA-\scriptsize{BLEU} & \cellcolor{cyan!50}31.38 & 45.79 & \cellcolor{cyan!50}38.14 & 25.64 & 41.25 & 32.09 & 31.95 & \cellcolor{red!50}27.57 & 40.05 & 29.37 & 31.39 \\
Ours-\scriptsize{BLEU} & \cellcolor{cyan!50}32.50 & 46.32 & \cellcolor{cyan!50}40.13 & 25.23 & 40.80 & 31.22 & 31.55 & \cellcolor{red!50}26.00 & 39.55 & 29.01 & 31.33 \\
ALMA-\scriptsize{COMET} & 85.57 & 87.71 & 87.82 & 81.38 & 86.26 & 86.84 & \cellcolor{red!50}90.90 & \cellcolor{red!50}87.61 & 87.14 & 88.80 & 78.12 \\
Ours-\scriptsize{COMET} & 85.50 & 87.67 & 87.71 & 81.24 & 86.17 & 86.02 & \cellcolor{red!50}89.84 & \cellcolor{red!50}85.80 & 86.39 & 87.89 & 77.63 \\
\cline{2-11}
\multicolumn{11}{c}{\textit{Hard instances}} \\
ALMA-\scriptsize{BLEU} & \cellcolor{cyan!50}12.25 & \cellcolor{cyan!50}29.49 & \cellcolor{cyan!50}21.72 & \cellcolor{cyan!50}1.95 & \cellcolor{cyan!50}15.73 & 15.71 & \cellcolor{cyan!50}12.79 & 17.51 & \cellcolor{cyan!50}14.59 & 15.45 & \cellcolor{cyan!50}14.17 \\
Ours-\scriptsize{BLEU} & \cellcolor{cyan!50}15.56 & \cellcolor{cyan!50}35.93 & \cellcolor{cyan!50}27.72 & \cellcolor{cyan!50}4.62 & \cellcolor{cyan!50}19.77 & 16.15 & \cellcolor{cyan!50}16.67 & 17.13 & \cellcolor{cyan!50}19.49 & 15.54 & \cellcolor{cyan!50}17.30 \\
ALMA-\scriptsize{COMET} & \cellcolor{cyan!50}62.73 & \cellcolor{cyan!50}67.08 & \cellcolor{cyan!50}72.62 & \cellcolor{cyan!50}49.94 & \cellcolor{cyan!50}62.64 & \cellcolor{cyan!50}58.50 & \cellcolor{cyan!50}60.80 & \cellcolor{cyan!50}70.02 & \cellcolor{cyan!50}59.07 & \cellcolor{cyan!50}62.31 & \cellcolor{cyan!50}56.34 \\
Ours-\scriptsize{COMET} & \cellcolor{cyan!50}65.98 & \cellcolor{cyan!50}71.16 & \cellcolor{cyan!50}75.12 & \cellcolor{cyan!50}58.99 & \cellcolor{cyan!50}67.19 & \cellcolor{cyan!50}60.90 & \cellcolor{cyan!50}67.90 & \cellcolor{cyan!50}71.57 & \cellcolor{cyan!50}62.03 & \cellcolor{cyan!50}65.16 & \cellcolor{cyan!50}60.08 \\
\hline
\multicolumn{11}{c}{\textit{N=200}} \\
\cline{2-11}
\multicolumn{11}{c}{\textit{Easy instances}} \\
ALMA-\scriptsize{BLEU} & \cellcolor{cyan!50}31.96 & 47.11 & 39.94 & 26.22 & 42.13 & 32.50 & 32.75 & \cellcolor{red!50}28.54 & 41.08 & 30.22 & 32.22 \\
Ours-\scriptsize{BLEU} & \cellcolor{cyan!50}33.10 & 47.41 & 41.60 & 25.79 & 41.43 & 31.52 & 32.20 & \cellcolor{red!50}26.91 & 40.48 & 29.79 & 32.04 \\
ALMA-\scriptsize{COMET} & 86.34 & 88.61 & 88.72 & 82.31 & 87.02 & \cellcolor{red!50}87.76 & \cellcolor{red!50}91.85 & \cellcolor{red!50}88.67 & 87.97 & 89.67 & 78.92 \\
Ours-\scriptsize{COMET} & 86.16 & 88.40 & 88.43 & 81.98 & 86.89 & \cellcolor{red!50}86.75 & \cellcolor{red!50}90.77 & \cellcolor{red!50}86.94 & 87.12 & 88.73 & 78.34 \\
\cline{2-11}
\multicolumn{11}{c}{\textit{Hard instances}} \\
ALMA-\scriptsize{BLEU} & \cellcolor{cyan!50}17.46 & \cellcolor{cyan!50}30.39 & \cellcolor{cyan!50}24.17 & \cellcolor{cyan!50}6.00 & \cellcolor{cyan!50}20.03 & 19.11 & \cellcolor{cyan!50}14.83 & 19.02 & \cellcolor{cyan!50}18.61 & 15.43 & \cellcolor{cyan!50}16.96 \\
Ours-\scriptsize{BLEU} & \cellcolor{cyan!50}19.31 & \cellcolor{cyan!50}35.04 & \cellcolor{cyan!50}29.25 & \cellcolor{cyan!50}7.55 & \cellcolor{cyan!50}23.70 & 19.96 & \cellcolor{cyan!50}18.16 & 18.29 & \cellcolor{cyan!50}21.52 & 15.95 & \cellcolor{cyan!50}19.28 \\
ALMA-\scriptsize{COMET} & \cellcolor{cyan!50}67.24 & \cellcolor{cyan!50}71.82 & \cellcolor{cyan!50}76.62 & \cellcolor{cyan!50}57.84 & \cellcolor{cyan!50}67.59 & \cellcolor{cyan!50}64.30 & \cellcolor{cyan!50}67.13 & 74.56 & 65.46 & \cellcolor{cyan!50}67.59 & \cellcolor{cyan!50}61.26 \\
Ours-\scriptsize{COMET} & \cellcolor{cyan!50}69.85 & \cellcolor{cyan!50}74.82 & \cellcolor{cyan!50}78.52 & \cellcolor{cyan!50}63.87 & \cellcolor{cyan!50}70.22 & \cellcolor{cyan!50}66.77 & \cellcolor{cyan!50}70.37 & 74.13 & 67.50 & \cellcolor{cyan!50}68.78 & \cellcolor{cyan!50}63.60 \\
\hline
\multicolumn{11}{c}{\textit{N=500}} \\
\cline{2-11}
\multicolumn{11}{c}{\textit{Easy instances}} \\
ALMA-\scriptsize{BLEU} & 34.36 & 50.81 & 46.92 & 28.50 & 45.16 & \cellcolor{red!50}34.61 & \cellcolor{red!50}35.28 & \cellcolor{red!50}31.79 & 43.91 & 32.13 & 35.13 \\
Ours-\scriptsize{BLEU} & 35.33 & 50.59 & 47.25 & 27.82 & 44.16 & \cellcolor{red!50}33.25 & \cellcolor{red!50}34.07 & \cellcolor{red!50}30.00 & 42.92 & 31.67 & 34.54 \\
ALMA-\scriptsize{COMET} & 88.08 & 90.54 & 91.04 & 84.29 & 88.62 & \cellcolor{red!50}89.59 & \cellcolor{red!50}93.66 & 91.08 & 89.79 & 91.47 & 80.67\\
Ours-\scriptsize{COMET} & 87.80 & 90.10 & 90.50 & 83.86 & 88.40 & \cellcolor{red!50}88.55 & \cellcolor{red!50}92.48 & 89.57 & 88.79 & 90.61 & 80.00\\
\cline{2-11}
\multicolumn{11}{c}{\textit{Hard instances}} \\
ALMA-\scriptsize{BLEU} & \cellcolor{cyan!50}21.31 & \cellcolor{cyan!50}35.46 & \cellcolor{cyan!50}28.66 & 13.08 & \cellcolor{cyan!50}25.4 & 22.53 & \cellcolor{cyan!50}19.82 & \cellcolor{red!50}22.52 & \cellcolor{cyan!50}24.81 & 19.78 & \cellcolor{cyan!50}21.36\\
Ours-\scriptsize{BLEU} & \cellcolor{cyan!50}23.09 & \cellcolor{cyan!50}37.91 & \cellcolor{cyan!50}32.66 & 14.04 & \cellcolor{cyan!50}27.32 & 22.89 & \cellcolor{cyan!50}22.38 & \cellcolor{red!50}21.32 & \cellcolor{cyan!50}26.58 & 19.78 & \cellcolor{cyan!50}22.82 \\
ALMA-\scriptsize{COMET} & \cellcolor{cyan!50}73.56 & \cellcolor{cyan!50}78.24 & 81.55 & \cellcolor{cyan!50}67.07 & \cellcolor{cyan!50}74.39 & 72.74 & 76.38 & \cellcolor{red!50}80.61 & 73.38 & 75.29 & 67.79 \\
Ours-\scriptsize{COMET} & \cellcolor{cyan!50}74.77 & \cellcolor{cyan!50}79.75 & 82.41 & \cellcolor{cyan!50}69.56 & \cellcolor{cyan!50}75.63 & 73.24 & 77.34 & \cellcolor{red!50}79.19 & 74.12 & 74.97 & 68.60 \\
\hline
\multicolumn{11}{c}{\textit{Overall performance, i.e., N=infinite when all instances are included.}} \\
ALMA-\scriptsize{BLEU} & \cellcolor{cyan!50}30.73 & 44.68 & \cellcolor{cyan!50}36.46 & 24.15 & 40.37 & 31.37 & 31.12 & \cellcolor{red!50}26.67 & 39.05 & 28.76 & 30.46 \\
Ours-\scriptsize{BLEU} & \cellcolor{cyan!50}31.93 & 45.60 & \cellcolor{cyan!50}38.85 & 23.94 & 40.09 & 30.64 & 30.91 & \cellcolor{red!50}25.22 & 38.76 & 28.43 & 30.59 \\
ALMA-\scriptsize{COMET} & 84.42 & 86.29 & 86.30 & 79.70 & 85.09 & 85.45 & 89.42 & \cellcolor{red!50}85.85 & 85.76 & 87.50 & 76.83 \\
Ours-\scriptsize{COMET} & 84.50 & 86.53 & 86.45 & 80.05 & 85.22 & 84.78 & 88.75 & \cellcolor{red!50}84.38 & 85.19 & 86.77 & 76.59\\
\hline
\end{tabular}
}
\caption{Specific results on 10 translation directions. The size of models are 13B. BLEU and COMET are reported. Cells where the difference is larger than $1.0$ are highlighted with colored background. \textcolor{cyan}{Blue} indicates ours model outperforms ALMA and \textcolor{red}{red} indicates the opposite.}
\label{tab:all_results}
\end{table*}

%% file: acl_latex.bbl
\begin{thebibliography}{51}
\expandafter\ifx\csname natexlab\endcsname\relax\def\natexlab#1{#1}\fi

\bibitem[{Achiam et~al.(2023)Achiam, Adler, Agarwal, Ahmad, Akkaya, Aleman, Almeida, Altenschmidt, Altman, Anadkat et~al.}]{achiam2023gpt}
Josh Achiam, Steven Adler, Sandhini Agarwal, Lama Ahmad, Ilge Akkaya, Florencia~Leoni Aleman, Diogo Almeida, Janko Altenschmidt, Sam Altman, Shyamal Anadkat, et~al. 2023.
\newblock Gpt-4 technical report.
\newblock \emph{arXiv preprint arXiv:2303.08774}.

\bibitem[{Alves et~al.(2024)Alves, Pombal, Guerreiro, Martins, Alves, Farajian, Peters, Rei, Fernandes, Agrawal, Colombo, de~Souza, and Martins}]{tower_llm_2024}
Duarte~M. Alves, José Pombal, Nuno~M. Guerreiro, Pedro~H. Martins, João Alves, Amin Farajian, Ben Peters, Ricardo Rei, Patrick Fernandes, Sweta Agrawal, Pierre Colombo, José G.~C. de~Souza, and André F.~T. Martins. 2024.
\newblock \href {http://arxiv.org/abs/2402.17733} {Tower: An open multilingual large language model for translation-related tasks}.

\bibitem[{Bahdanau et~al.(2015)Bahdanau, Cho, and Bengio}]{BahdanauCB14}
Dzmitry Bahdanau, Kyunghyun Cho, and Yoshua Bengio. 2015.
\newblock \href {http://arxiv.org/abs/1409.0473} {Neural machine translation by jointly learning to align and translate}.
\newblock In \emph{3rd International Conference on Learning Representations, {ICLR} 2015, San Diego, CA, USA, May 7-9, 2015, Conference Track Proceedings}.

\bibitem[{Bang et~al.(2023)Bang, Cahyawijaya, Lee, Dai, Su, Wilie, Lovenia, Ji, Yu, Chung et~al.}]{Bang2023AMM}
Yejin Bang, Samuel Cahyawijaya, Nayeon Lee, Wenliang Dai, Dan Su, Bryan Wilie, Holy Lovenia, Ziwei Ji, Tiezheng Yu, Willy Chung, et~al. 2023.
\newblock A multitask, multilingual, multimodal evaluation of chatgpt on reasoning, hallucination, and interactivity.
\newblock In \emph{Proceedings of the 13th International Joint Conference on Natural Language Processing and the 3rd Conference of the Asia-Pacific Chapter of the Association for Computational Linguistics (Volume 1: Long Papers)}, pages 675--718.

\bibitem[{Barrault et~al.(2020)Barrault, Biesialska, Bojar, Costa-juss{\`a}, Federmann, Graham, Grundkiewicz, Haddow, Huck, Joanis, Kocmi, Koehn, Lo, Ljube{\v{s}}i{\'c}, Monz, Morishita, Nagata, Nakazawa, Pal, Post, and Zampieri}]{barrault-etal-2020-findings}
Lo{\"\i}c Barrault, Magdalena Biesialska, Ond{\v{r}}ej Bojar, Marta~R. Costa-juss{\`a}, Christian Federmann, Yvette Graham, Roman Grundkiewicz, Barry Haddow, Matthias Huck, Eric Joanis, Tom Kocmi, Philipp Koehn, Chi-kiu Lo, Nikola Ljube{\v{s}}i{\'c}, Christof Monz, Makoto Morishita, Masaaki Nagata, Toshiaki Nakazawa, Santanu Pal, Matt Post, and Marcos Zampieri. 2020.
\newblock \href {https://aclanthology.org/2020.wmt-1.1} {Findings of the 2020 conference on machine translation ({WMT}20)}.
\newblock In \emph{Proceedings of the Fifth Conference on Machine Translation}, pages 1--55, Online. Association for Computational Linguistics.

\bibitem[{Bojar et~al.(2017)Bojar, Chatterjee, Federmann, Graham, Haddow, Huang, Huck, Koehn, Liu, Logacheva, Monz, Negri, Post, Rubino, Specia, and Turchi}]{bojar-etal-2017-findings}
Ond{\v{r}}ej Bojar, Rajen Chatterjee, Christian Federmann, Yvette Graham, Barry Haddow, Shujian Huang, Matthias Huck, Philipp Koehn, Qun Liu, Varvara Logacheva, Christof Monz, Matteo Negri, Matt Post, Raphael Rubino, Lucia Specia, and Marco Turchi. 2017.
\newblock \href {https://doi.org/10.18653/v1/W17-4717} {Findings of the 2017 conference on machine translation ({WMT}17)}.
\newblock In \emph{Proceedings of the Second Conference on Machine Translation}, pages 169--214, Copenhagen, Denmark. Association for Computational Linguistics.

\bibitem[{Brown et~al.(2020)Brown, Mann, Ryder, Subbiah, Kaplan, Dhariwal, Neelakantan, Shyam, Sastry, Askell, Agarwal, Herbert-Voss, Krueger, Henighan, Child, Ramesh, Ziegler, Wu, Winter, Hesse, Chen, Sigler, Litwin, Gray, Chess, Clark, Berner, McCandlish, Radford, Sutskever, and Amodei}]{gpt3}
Tom Brown, Benjamin Mann, Nick Ryder, Melanie Subbiah, Jared~D Kaplan, Prafulla Dhariwal, Arvind Neelakantan, Pranav Shyam, Girish Sastry, Amanda Askell, Sandhini Agarwal, Ariel Herbert-Voss, Gretchen Krueger, Tom Henighan, Rewon Child, Aditya Ramesh, Daniel Ziegler, Jeffrey Wu, Clemens Winter, Chris Hesse, Mark Chen, Eric Sigler, Mateusz Litwin, Scott Gray, Benjamin Chess, Jack Clark, Christopher Berner, Sam McCandlish, Alec Radford, Ilya Sutskever, and Dario Amodei. 2020.
\newblock \href {https://proceedings.neurips.cc/paper_files/paper/2020/file/1457c0d6bfcb4967418bfb8ac142f64a-Paper.pdf} {Language models are few-shot learners}.
\newblock In \emph{Advances in Neural Information Processing Systems}, volume~33, pages 1877--1901. Curran Associates, Inc.

\bibitem[{Chi et~al.(2021)Chi, Dong, Zheng, Huang, Mao, Huang, and Wei}]{chi-etal-2021-improving}
Zewen Chi, Li~Dong, Bo~Zheng, Shaohan Huang, Xian-Ling Mao, Heyan Huang, and Furu Wei. 2021.
\newblock \href {https://doi.org/10.18653/v1/2021.acl-long.265} {Improving pretrained cross-lingual language models via self-labeled word alignment}.
\newblock In \emph{Proceedings of the 59th Annual Meeting of the Association for Computational Linguistics and the 11th International Joint Conference on Natural Language Processing (Volume 1: Long Papers)}, pages 3418--3430, Online. Association for Computational Linguistics.

\bibitem[{Chousa et~al.(2020)Chousa, Nagata, and Nishino}]{chousa-etal-2020-spanalign}
Katsuki Chousa, Masaaki Nagata, and Masaaki Nishino. 2020.
\newblock \href {https://doi.org/10.18653/v1/2020.coling-main.418} {{S}pan{A}lign: Sentence alignment method based on cross-language span prediction and {ILP}}.
\newblock In \emph{Proceedings of the 28th International Conference on Computational Linguistics}, pages 4750--4761, Barcelona, Spain (Online). International Committee on Computational Linguistics.

\bibitem[{Chung et~al.(2024)Chung, Hou, Longpre, Zoph, Tay, Fedus, Li, Wang, Dehghani, Brahma et~al.}]{flan-t5}
Hyung~Won Chung, Le~Hou, Shayne Longpre, Barret Zoph, Yi~Tay, William Fedus, Yunxuan Li, Xuezhi Wang, Mostafa Dehghani, Siddhartha Brahma, et~al. 2024.
\newblock Scaling instruction-finetuned language models.
\newblock \emph{Journal of Machine Learning Research}, 25(70):1--53.

\bibitem[{Costa-juss{\`a} et~al.(2022)Costa-juss{\`a}, Cross, {\c{C}}elebi, Elbayad, Heafield, Heffernan, Kalbassi, Lam, Licht, Maillard et~al.}]{nllb2022}
Marta~R Costa-juss{\`a}, James Cross, Onur {\c{C}}elebi, Maha Elbayad, Kenneth Heafield, Kevin Heffernan, Elahe Kalbassi, Janice Lam, Daniel Licht, Jean Maillard, et~al. 2022.
\newblock No language left behind: Scaling human-centered machine translation.
\newblock \emph{arXiv preprint arXiv:2207.04672}.

\bibitem[{Dale et~al.(2023{\natexlab{a}})Dale, Voita, Barrault, and Costa-juss{\`a}}]{dale-etal-2023-detecting}
David Dale, Elena Voita, Loic Barrault, and Marta~R. Costa-juss{\`a}. 2023{\natexlab{a}}.
\newblock \href {https://doi.org/10.18653/v1/2023.acl-long.3} {Detecting and mitigating hallucinations in machine translation: Model internal workings alone do well, sentence similarity {E}ven better}.
\newblock In \emph{Proceedings of the 61st Annual Meeting of the Association for Computational Linguistics (Volume 1: Long Papers)}, pages 36--50, Toronto, Canada. Association for Computational Linguistics.

\bibitem[{Dale et~al.(2023{\natexlab{b}})Dale, Voita, Lam, Hansanti, Ropers, Kalbassi, Gao, Barrault, and Costa-juss{\`a}}]{dale-etal-2023-halomi}
David Dale, Elena Voita, Janice Lam, Prangthip Hansanti, Christophe Ropers, Elahe Kalbassi, Cynthia Gao, Loic Barrault, and Marta Costa-juss{\`a}. 2023{\natexlab{b}}.
\newblock \href {https://doi.org/10.18653/v1/2023.emnlp-main.42} {{H}al{O}mi: A manually annotated benchmark for multilingual hallucination and omission detection in machine translation}.
\newblock In \emph{Proceedings of the 2023 Conference on Empirical Methods in Natural Language Processing}, pages 638--653, Singapore. Association for Computational Linguistics.

\bibitem[{Dhuliawala et~al.(2023)Dhuliawala, Komeili, Xu, Raileanu, Li, Celikyilmaz, and Weston}]{Dhuliawala2023ChainofVerificationRH}
Shehzaad Dhuliawala, Mojtaba Komeili, Jing Xu, Roberta Raileanu, Xian Li, Asli Celikyilmaz, and Jason Weston. 2023.
\newblock \href {https://api.semanticscholar.org/CorpusID:262062565} {Chain-of-verification reduces hallucination in large language models}.
\newblock \emph{ArXiv}, abs/2309.11495.

\bibitem[{Dou and Neubig(2021)}]{dou2021word}
Zi-Yi Dou and Graham Neubig. 2021.
\newblock Word alignment by fine-tuning embeddings on parallel corpora.
\newblock In \emph{Proceedings of the 16th Conference of the European Chapter of the Association for Computational Linguistics: Main Volume}, pages 2112--2128.

\bibitem[{Dyer et~al.(2013)Dyer, Chahuneau, and Smith}]{dyer2013simple}
Chris Dyer, Victor Chahuneau, and Noah~A Smith. 2013.
\newblock A simple, fast, and effective reparameterization of ibm model 2.
\newblock In \emph{Proceedings of the 2013 Conference of the North American Chapter of the Association for Computational Linguistics: Human Language Technologies}, pages 644--648.

\bibitem[{Feng et~al.(2022)Feng, Yang, Cer, Arivazhagan, and Wang}]{feng-etal-2022-language}
Fangxiaoyu Feng, Yinfei Yang, Daniel Cer, Naveen Arivazhagan, and Wei Wang. 2022.
\newblock \href {https://doi.org/10.18653/v1/2022.acl-long.62} {Language-agnostic {BERT} sentence embedding}.
\newblock In \emph{Proceedings of the 60th Annual Meeting of the Association for Computational Linguistics (Volume 1: Long Papers)}, pages 878--891, Dublin, Ireland. Association for Computational Linguistics.

\bibitem[{Gao et~al.(2021)Gao, Yao, and Chen}]{gao-etal-2021-simcse}
Tianyu Gao, Xingcheng Yao, and Danqi Chen. 2021.
\newblock \href {https://doi.org/10.18653/v1/2021.emnlp-main.552} {{S}im{CSE}: Simple contrastive learning of sentence embeddings}.
\newblock In \emph{Proceedings of the 2021 Conference on Empirical Methods in Natural Language Processing}, pages 6894--6910, Online and Punta Cana, Dominican Republic. Association for Computational Linguistics.

\bibitem[{Hendy et~al.(2023)Hendy, Abdelrehim, Sharaf, Raunak, Gabr, Matsushita, Kim, Afify, and Awadalla}]{gpt-mt-2023}
Amr Hendy, Mohamed Abdelrehim, Amr Sharaf, Vikas Raunak, Mohamed Gabr, Hitokazu Matsushita, Young~Jin Kim, Mohamed Afify, and Hany~Hassan Awadalla. 2023.
\newblock How good are gpt models at machine translation? a comprehensive evaluation.
\newblock \emph{arXiv preprint arXiv:2302.09210}.

\bibitem[{Hu et~al.(2021)Hu, Wallis, Allen-Zhu, Li, Wang, Wang, Chen et~al.}]{Hu2021LoRALA}
Edward~J Hu, Phillip Wallis, Zeyuan Allen-Zhu, Yuanzhi Li, Shean Wang, Lu~Wang, Weizhu Chen, et~al. 2021.
\newblock Lora: Low-rank adaptation of large language models.
\newblock In \emph{International Conference on Learning Representations}.

\bibitem[{Jalili~Sabet et~al.(2020)Jalili~Sabet, Dufter, Yvon, and Sch{\"u}tze}]{JaliliSabet2020SimAlignHQ}
Masoud Jalili~Sabet, Philipp Dufter, Fran{\c{c}}ois Yvon, and Hinrich Sch{\"u}tze. 2020.
\newblock {S}im{A}lign: High quality word alignments without parallel training data using static and contextualized embeddings.
\newblock In \emph{Findings of the Association for Computational Linguistics: EMNLP 2020}, pages 1627--1643, Online. Association for Computational Linguistics.

\bibitem[{Kocmi et~al.(2024)Kocmi, Avramidis, Bawden, Bojar, Dvorkovich, Federmann, Fishel, Freitag, Gowda, Grundkiewicz et~al.}]{kocmi2024preliminary}
Tom Kocmi, Eleftherios Avramidis, Rachel Bawden, Ondrej Bojar, Anton Dvorkovich, Christian Federmann, Mark Fishel, Markus Freitag, Thamme Gowda, Roman Grundkiewicz, et~al. 2024.
\newblock Preliminary wmt24 ranking of general mt systems and llms.
\newblock \emph{arXiv preprint arXiv:2407.19884}.

\bibitem[{Kojima et~al.(2022)Kojima, Gu, Reid, Matsuo, and Iwasawa}]{Kojima2022LargeLM}
Takeshi Kojima, Shixiang~Shane Gu, Machel Reid, Yutaka Matsuo, and Yusuke Iwasawa. 2022.
\newblock Large language models are zero-shot reasoners.
\newblock \emph{Advances in neural information processing systems}, 35:22199--22213.

\bibitem[{Li et~al.(2023)Li, Huang, Zhang, Deng, Lou, Huang, Jiao, Wei, Deng, and Zhang}]{li-etal-2023-dual}
Ziheng Li, Shaohan Huang, Zihan Zhang, Zhi-Hong Deng, Qiang Lou, Haizhen Huang, Jian Jiao, Furu Wei, Weiwei Deng, and Qi~Zhang. 2023.
\newblock \href {https://doi.org/10.18653/v1/2023.acl-long.191} {Dual-alignment pre-training for cross-lingual sentence embedding}.
\newblock In \emph{Proceedings of the 61st Annual Meeting of the Association for Computational Linguistics (Volume 1: Long Papers)}, pages 3466--3478, Toronto, Canada. Association for Computational Linguistics.

\bibitem[{Miao et~al.(2024)Miao, Wu, Zhao, Wu, and Tsuruoka}]{miao2024enhancing}
Zhongtao Miao, Qiyu Wu, Kaiyan Zhao, Zilong Wu, and Yoshimasa Tsuruoka. 2024.
\newblock Enhancing cross-lingual sentence embedding for low-resource languages with word alignment.
\newblock \emph{arXiv preprint arXiv:2404.02490}.

\bibitem[{Mitchell et~al.(2023)Mitchell, Lee, Khazatsky, Manning, and Finn}]{Mitchell2023DetectGPTZM}
Eric Mitchell, Yoonho Lee, Alexander Khazatsky, Christopher~D. Manning, and Chelsea Finn. 2023.
\newblock \href {https://api.semanticscholar.org/CorpusID:256274849} {Detectgpt: Zero-shot machine-generated text detection using probability curvature}.
\newblock In \emph{International Conference on Machine Learning}.

\bibitem[{Nagata et~al.(2020)Nagata, Chousa, and Nishino}]{nagata2020supervised}
Masaaki Nagata, Katsuki Chousa, and Masaaki Nishino. 2020.
\newblock A supervised word alignment method based on cross-language span prediction using multilingual bert.
\newblock In \emph{Proceedings of the 2020 Conference on Empirical Methods in Natural Language Processing (EMNLP)}, pages 555--565.

\bibitem[{Och and Ney(2003)}]{och2003systematic}
Franz~Josef Och and Hermann Ney. 2003.
\newblock A systematic comparison of various statistical alignment models.
\newblock \emph{Computational linguistics}, 29(1):19--51.

\bibitem[{Ouyang et~al.(2022)Ouyang, Wu, Jiang, Almeida, Wainwright, Mishkin, Zhang, Agarwal, Slama, Ray, Schulman, Hilton, Kelton, Miller, Simens, Askell, Welinder, Christiano, Leike, and Lowe}]{instruct_gpt}
Long Ouyang, Jeffrey Wu, Xu~Jiang, Diogo Almeida, Carroll Wainwright, Pamela Mishkin, Chong Zhang, Sandhini Agarwal, Katarina Slama, Alex Ray, John Schulman, Jacob Hilton, Fraser Kelton, Luke Miller, Maddie Simens, Amanda Askell, Peter Welinder, Paul~F Christiano, Jan Leike, and Ryan Lowe. 2022.
\newblock \href {https://proceedings.neurips.cc/paper_files/paper/2022/file/b1efde53be364a73914f58805a001731-Paper-Conference.pdf} {Training language models to follow instructions with human feedback}.
\newblock In \emph{Advances in Neural Information Processing Systems}, volume~35, pages 27730--27744. Curran Associates, Inc.

\bibitem[{Papineni et~al.(2002)Papineni, Roukos, Ward, and Zhu}]{papineni-etal-2002-bleu}
Kishore Papineni, Salim Roukos, Todd Ward, and Wei-Jing Zhu. 2002.
\newblock \href {https://doi.org/10.3115/1073083.1073135} {{B}leu: a method for automatic evaluation of machine translation}.
\newblock In \emph{Proceedings of the 40th Annual Meeting of the Association for Computational Linguistics}, pages 311--318, Philadelphia, Pennsylvania, USA. Association for Computational Linguistics.

\bibitem[{Peng et~al.(2023)Peng, Wang, Dong, Hao, Huang, Ma, and Wei}]{peng2023kosmos}
Zhiliang Peng, Wenhui Wang, Li~Dong, Yaru Hao, Shaohan Huang, Shuming Ma, and Furu Wei. 2023.
\newblock Kosmos-2: Grounding multimodal large language models to the world.
\newblock \emph{arXiv preprint arXiv:2306.14824}.

\bibitem[{Radford et~al.(2019)Radford, Wu, Child, Luan, Amodei, Sutskever et~al.}]{gpt2}
Alec Radford, Jeffrey Wu, Rewon Child, David Luan, Dario Amodei, Ilya Sutskever, et~al. 2019.
\newblock Language models are unsupervised multitask learners.
\newblock \emph{OpenAI blog}, 1(8):9.

\bibitem[{Rafailov et~al.(2024)Rafailov, Sharma, Mitchell, Manning, Ermon, and Finn}]{dpo}
Rafael Rafailov, Archit Sharma, Eric Mitchell, Christopher~D Manning, Stefano Ermon, and Chelsea Finn. 2024.
\newblock Direct preference optimization: Your language model is secretly a reward model.
\newblock \emph{Advances in Neural Information Processing Systems}, 36.

\bibitem[{Touvron et~al.(2023)Touvron, Martin, Stone, Albert, Almahairi, Babaei, Bashlykov, Batra, Bhargava, Bhosale et~al.}]{touvron2023llama}
Hugo Touvron, Louis Martin, Kevin Stone, Peter Albert, Amjad Almahairi, Yasmine Babaei, Nikolay Bashlykov, Soumya Batra, Prajjwal Bhargava, Shruti Bhosale, et~al. 2023.
\newblock Llama 2: Open foundation and fine-tuned chat models.
\newblock \emph{arXiv preprint arXiv:2307.09288}.

\bibitem[{Tu et~al.(2016)Tu, Lu, Liu, Liu, and Li}]{Tu2016ModelingCF}
Zhaopeng Tu, Zhengdong Lu, Yang Liu, Xiaohua Liu, and Hang Li. 2016.
\newblock Modeling coverage for neural machine translation.
\newblock In \emph{Proceedings of the 54th Annual Meeting of the Association for Computational Linguistics (Volume 1: Long Papers)}, pages 76--85.

\bibitem[{Tunstall et~al.(2023)Tunstall, Beeching, Lambert, Rajani, Rasul, Belkada, Huang, von Werra, Fourrier, Habib, Sarrazin, Sanseviero, Rush, and Wolf}]{tunstall2023zephyr}
Lewis Tunstall, Edward Beeching, Nathan Lambert, Nazneen Rajani, Kashif Rasul, Younes Belkada, Shengyi Huang, Leandro von Werra, Cl{\'e}mentine Fourrier, Nathan Habib, Nathan Sarrazin, Omar Sanseviero, Alexander~M. Rush, and Thomas Wolf. 2023.
\newblock \href {https://api.semanticscholar.org/CorpusID:264490502} {Zephyr: Direct distillation of lm alignment}.
\newblock \emph{ArXiv}, abs/2310.16944.

\bibitem[{Vamvas and Sennrich(2022)}]{vamvas-sennrich-2022-little}
Jannis Vamvas and Rico Sennrich. 2022.
\newblock \href {https://doi.org/10.18653/v1/2022.acl-short.53} {As little as possible, as much as necessary: Detecting over- and undertranslations with contrastive conditioning}.
\newblock In \emph{Proceedings of the 60th Annual Meeting of the Association for Computational Linguistics (Volume 2: Short Papers)}, pages 490--500, Dublin, Ireland. Association for Computational Linguistics.

\bibitem[{Wei et~al.(2022)Wei, Bosma, Zhao, Guu, Yu, Lester, Du, Dai, and Le}]{flan}
Jason Wei, Maarten Bosma, Vincent Zhao, Kelvin Guu, Adams~Wei Yu, Brian Lester, Nan Du, Andrew~M. Dai, and Quoc~V Le. 2022.
\newblock \href {https://openreview.net/forum?id=gEZrGCozdqR} {Finetuned language models are zero-shot learners}.
\newblock In \emph{International Conference on Learning Representations}.

\bibitem[{Wei et~al.(2023)Wei, Cui, Cheng, Wang, Zhang, Huang, Xie, Xu, Chen, Zhang, Jiang, and Han}]{Wei2023ZeroShotIE}
Xiang Wei, Xingyu Cui, Ning Cheng, Xiaobin Wang, Xin Zhang, Shen Huang, Pengjun Xie, Jinan Xu, Yufeng Chen, Meishan Zhang, Yong Jiang, and Wenjuan Han. 2023.
\newblock \href {https://api.semanticscholar.org/CorpusID:257050669} {Zero-shot information extraction via chatting with chatgpt}.
\newblock \emph{ArXiv}, abs/2302.10205.

\bibitem[{Wu et~al.(2023{\natexlab{a}})Wu, Nagata, and Tsuruoka}]{wu-etal-2023-wspalign}
Qiyu Wu, Masaaki Nagata, and Yoshimasa Tsuruoka. 2023{\natexlab{a}}.
\newblock \href {https://doi.org/10.18653/v1/2023.acl-long.621} {{WSPA}lign: Word alignment pre-training via large-scale weakly supervised span prediction}.
\newblock In \emph{Proceedings of the 61st Annual Meeting of the Association for Computational Linguistics (Volume 1: Long Papers)}, pages 11084--11099, Toronto, Canada. Association for Computational Linguistics.

\bibitem[{Wu et~al.(2022)Wu, Tao, Shen, Xu, Geng, and Jiang}]{wu-etal-2022-pcl}
Qiyu Wu, Chongyang Tao, Tao Shen, Can Xu, Xiubo Geng, and Daxin Jiang. 2022.
\newblock \href {https://doi.org/10.18653/v1/2022.emnlp-main.826} {{PCL}: Peer-contrastive learning with diverse augmentations for unsupervised sentence embeddings}.
\newblock In \emph{Proceedings of the 2022 Conference on Empirical Methods in Natural Language Processing}, pages 12052--12066, Abu Dhabi, United Arab Emirates. Association for Computational Linguistics.

\bibitem[{Wu et~al.(2024)Wu, Wu, and Tsuruoka}]{wu2024sga}
Qiyu Wu, Zilong Wu, and Yoshimasa Tsuruoka. 2024.
\newblock \href {https://human-interpretable-ai.github.io/assets/pdf/10_SGA_Scene_Graph_Alignment_f.pdf} {Sga: Scene graph alignment for evaluation of text-to-image generation}.
\newblock \emph{HI-AI@KDD, Human-Interpretable AI Workshop at the KDD 2024, Barcelona, Spain}.

\bibitem[{Wu et~al.(2021)Wu, Xing, Li, Ke, He, and Liu}]{Wu2021TakingNO}
Qiyu Wu, Chen Xing, Yatao Li, Guolin Ke, Di~He, and Tie-Yan Liu. 2021.
\newblock \href {https://api.semanticscholar.org/CorpusID:235613669} {Taking notes on the fly helps language pre-training}.
\newblock In \emph{International Conference on Learning Representations}.

\bibitem[{Wu et~al.(2023{\natexlab{b}})Wu, Zhao, He, Huang, Ono, Wakaki, and Mitsufuji}]{wu2023towards}
Qiyu Wu, Mengjie Zhao, Yutong He, Lang Huang, Junya Ono, Hiromi Wakaki, and Yuki Mitsufuji. 2023{\natexlab{b}}.
\newblock Towards reporting bias in visual-language datasets: bimodal augmentation by decoupling object-attribute association.
\newblock \emph{arXiv preprint arXiv:2310.01330}.

\bibitem[{Xie et~al.(2022)Xie, Wu, Chen, and Wang}]{xie2022stable}
Yutao Xie, Qiyu Wu, Wei Chen, and Tengjiao Wang. 2022.
\newblock Stable contrastive learning for self-supervised sentence embeddings with pseudo-siamese mutual learning.
\newblock \emph{IEEE/ACM Transactions on Audio, Speech, and Language Processing}, 30:3046--3059.

\bibitem[{Xu et~al.(2024{\natexlab{a}})Xu, Kim, Sharaf, and Awadalla}]{alma}
Haoran Xu, Young~Jin Kim, Amr Sharaf, and Hany~Hassan Awadalla. 2024{\natexlab{a}}.
\newblock \href {https://openreview.net/forum?id=farT6XXntP} {A paradigm shift in machine translation: Boosting translation performance of large language models}.
\newblock In \emph{The Twelfth International Conference on Learning Representations}.

\bibitem[{Xu et~al.(2024{\natexlab{b}})Xu, Sharaf, Chen, Tan, Shen, Van~Durme, Murray, and Kim}]{cpo}
Haoran Xu, Amr Sharaf, Yunmo Chen, Weiting Tan, Lingfeng Shen, Benjamin Van~Durme, Kenton Murray, and Young~Jin Kim. 2024{\natexlab{b}}.
\newblock Contrastive preference optimization: Pushing the boundaries of llm performance in machine translation.
\newblock \emph{arXiv preprint arXiv:2401.08417}.

\bibitem[{Yang et~al.(2019)Yang, Cheng, Liu, and Sun}]{yang-etal-2019-reducing}
Zonghan Yang, Yong Cheng, Yang Liu, and Maosong Sun. 2019.
\newblock \href {https://doi.org/10.18653/v1/P19-1623} {Reducing word omission errors in neural machine translation: A contrastive learning approach}.
\newblock In \emph{Proceedings of the 57th Annual Meeting of the Association for Computational Linguistics}, pages 6191--6196, Florence, Italy. Association for Computational Linguistics.

\bibitem[{Zhang et~al.(2023{\natexlab{a}})Zhang, Li, Cui, Cai, Liu, Fu, Huang, Zhao, Zhang, Chen, Wang, Luu, Bi, Shi, and Shi}]{Zhang2023SirensSI}
Yue Zhang, Yafu Li, Leyang Cui, Deng Cai, Lemao Liu, Tingchen Fu, Xinting Huang, Enbo Zhao, Yu~Zhang, Yulong Chen, Longyue Wang, Anh~Tuan Luu, Wei Bi, Freda Shi, and Shuming Shi. 2023{\natexlab{a}}.
\newblock \href {https://api.semanticscholar.org/CorpusID:261530162} {Siren's song in the ai ocean: A survey on hallucination in large language models}.
\newblock \emph{ArXiv}, abs/2309.01219.

\bibitem[{Zhang et~al.(2023{\natexlab{b}})Zhang, Tan, Huang, and Huang}]{zhang2023veco}
Zhen-Ru Zhang, Chuanqi Tan, Songfang Huang, and Fei Huang. 2023{\natexlab{b}}.
\newblock Veco 2.0: Cross-lingual language model pre-training with multi-granularity contrastive learning.
\newblock \emph{arXiv preprint arXiv:2304.08205}.

\bibitem[{Zhao et~al.(2024)Zhao, Wu, Cai, and Tsuruoka}]{zhao-etal-2024-leveraging}
Kaiyan Zhao, Qiyu Wu, Xin-Qiang Cai, and Yoshimasa Tsuruoka. 2024.
\newblock \href {https://aclanthology.org/2024.eacl-long.59} {Leveraging multi-lingual positive instances in contrastive learning to improve sentence embedding}.
\newblock In \emph{Proceedings of the 18th Conference of the European Chapter of the Association for Computational Linguistics (Volume 1: Long Papers)}, pages 976--991, St. Julian{'}s, Malta. Association for Computational Linguistics.

\end{thebibliography}
